\documentclass[oneside]{article}
\usepackage[accepted]{tmlr}
\usepackage[perpage]{footmisc}
\usepackage{amsmath}
\usepackage{amsfonts}
\usepackage{dsfont}
\usepackage{fontenc}
\usepackage{booktabs} 
\usepackage{caption} 
\usepackage{subfigure} 
\usepackage{graphicx}
\usepackage{enumerate}
\usepackage{stmaryrd}
\usepackage{pgfplots}
\usepackage[all]{nowidow}
\usepackage[utf8]{inputenc}
\usepackage{tikz}
\usepackage{tikz-cd}
\usetikzlibrary{babel}
\usetikzlibrary{er,positioning,bayesnet}
\usepackage{multicol}
\usepackage{algorithm2e}
\usepackage{hyperref}
\usepackage[linecolor=blue!60!,backgroundcolor=blue!10!,textwidth=25mm,textsize=scriptsize]{todonotes}

\definecolor{blue}{HTML}{1F77B4}
\definecolor{orange}{HTML}{FF7F0E}
\definecolor{green}{HTML}{2CA02C}

\pgfplotsset{compat=1.14}

\newtheorem{proposition}{Proposition}

\begin{document}
\title{Hypergraph clustering using Ricci curvature: an edge transport perspective}
%
%
\author{\name Olympio Hacquard \email hacquard.olympio.4w@kyoto-u.ac.jp \\
       \addr Kyoto University Institute for Advanced Study\\
       Kyoto University, Kyoto 606-8501, Japan
       }
       
\def\month{10}  
\def\year{2025} 
\def\openreview{\url{https://openreview.net/forum?id=HMROU8MXqV}} 

%
%
%

\maketitle              
\begin{abstract}
In this paper, we introduce a novel method for extending Ricci flow to hypergraphs by defining probability measures on the edges and transporting them on the line expansion. This approach yields a new weighting on the edges, which proves particularly effective for community detection. We extensively compare this method with a similar notion of Ricci flow defined on the clique expansion, demonstrating its enhanced sensitivity to the hypergraph structure, especially in the presence of large hyperedges. The two methods are complementary and together form a powerful and highly interpretable framework for community detection in hypergraphs.
\end{abstract}


\section{Introduction}

Hypergraphs are generalizations of graphs extending the concept of edges to encompass relationships involving any number of vertices rather than being restricted to pairs.  Hypergraphs provide a more faithful data representation than graphs as many real-life interactions occur at a group level rather than at a binary one. Examples include social interactions, simultaneous participation at an event, co-authorship, mutual evolutionary traits, spatial proximity or molecular interactions. These interactions are better encoded by hypergraphs instead of traditional graphs, see \cite{davis1941south, dotson2022deciphering, konstantinova2001application, torres2021and, klamt2009hypergraphs}. We refer to \cite{berge1984hypergraphs} and \cite{bretto2013hypergraph} for a comprehensive overview of hypergraph theory. 

Recently, machine learning tasks on hypergraphs have encountered a growing success, in particular for classification \citep{feng2019hypergraph}, regression \citep{liu2021persistent}, anomaly detection \citep{lee2022hashnwalk} or embedding \citep{zhang2019hyper, antelmi2023survey}. The present article focuses on the clustering problem or community detection, i.e. recovering labels on the nodes in a fully unsupervised setting from a single hypergraph instance. Several approaches have been taken to tackle this problem such as embeddings using neural networks \citep{zhang2021double, lee2023m}, random walks-based embeddings \citep{hayashi2020hypergraph, huang2019hyper2vec} followed by a clustering in the Euclidean space. Some methods developed in \cite{kumar2020new, kaminski2024modularity} aim at maximizing hypergraph-modularity functions that measure the strength of a clustering. Finally, let us mention the work of \cite{eyubov2023freight} that finds a partition in linear time by streaming all the nodes one by one.

The present work investigates generalizations of Ricci flow based clustering on hypergraphs. Ollivier-Ricci curvature, introduced by \cite{ollivier2007ricci} for general metric spaces and later specifically tailored to graphs \citep{lin2011ricci}, defines a distance on graph edges that quantifies local curvature using optimal transport theory. Edges within strongly connected communities have a positive curvature while edges bridging two communities are negatively curved. Ricci curvature can be turned into a flow dynamic in order to further stress out this community structure and take some more global graph properties into account. Alternative notions of curvature for graph processing are also present in \cite{forman2003bochner, sreejith2016forman, iyer2024non, tian2025curvature}. Ricci curvature on graphs has been used to derive theoretical bounds on the spectrum of the graph Laplacian by \cite{lin2010ricci} and \cite{bauer2011ollivier}. For applied purposes, it has been used for data exploration \citep{ni2015ricci}, representation learning \citep{zhang2023ricci}, topological data analysis \citep{carriere2020multiparameter, hacquard2024euler}, and in particular clustering \citep{ni2019community}, relying on the property of the Ricci flow to place emphasis on the community structure.

There has been a few attempts to generalize Ollivier-Ricci curvature to hypergraphs, in particular by \cite{coupette2022ollivier}. The authors introduce a notion of curvature using random walks on the nodes. As pointed out by \cite{chitra2019random}, in most practical cases, considering a random walk on the hypergraph is equivalent to a random walk on its \textit{clique expansion} (a graph representation where each hyperedge is replaced by a weighted clique). Doing so is a common way to circumvent the complex hypergraph structure by replacing it with a graph with a similar structure. However, as noted in \cite{chitra2019random}, this reduction can result in significant information loss, as distinct hypergraphs may share identical clique graphs \citep{hein2013total}. In addition, replacing a single hyperedge between $k$ nodes implies adding up to ${k \choose 2}$ connections between nodes which can be computationally prohibitive in a network with mutual interactions between many agents. Extensions of Ricci curvature on directed hypergraphs have also been proposed by \cite{eidi2020ollivier}.

The main contribution of this work aims at defining a notion of Ricci curvature on hypergraphs where we consider probability distributions on the edges instead than on the nodes. This approach leverages the \textit{line expansion} of the hypergraph, which is a graph representation where nodes correspond to hyperedges, and edges reflect hyperedges intersections. The line expansion is a common way to represent a hypergraph for learning purposes, see \cite{bandyopadhyay2020hypergraph, yang2022semi}. In addition, it has been demonstrated in \cite{kirkland2018two} that almost all the hypergraph information is retained by the joint knowledge of both its clique and line expansions. The current work puts a particular focus on clustering. We perform a thorough experimental study on synthetic and real data comparing the approach where we transport measures on the edges to the more standard approach where we transport measures on the nodes. More precisely, this notion of edge transport should be favored when we have a large number of small communities, when the edges between communities have a larger cardinality than the ones within communities, and more generally for hypergraphs with very large hyperedges as this approach is much more efficient computationally.

The work is organized as follows: Section \ref{sec:model} introduces the foundational concepts of hypergraph theory and Ollivier-Ricci curvature on graphs. Section \ref{sec:methods} presents two different possible expansions of Ricci curvature for hypergraphs: the standard one where we transport measures on the nodes, and the main contribution of this paper where we transport measures on the edges. We also compare the two approaches theoretically on a synthetic example. Section \ref{sec:expe} provides a detailed baseline of experiments on synthetic data and on real data where we compare both methods to state-of-the-art clustering algorithms on hypergraphs, along with a computational analysis of both methods.

\section{Model}
\label{sec:model}

This section presents the main concepts related to hypergraph analysis and the hypergraph partitioning problem. We also investigate standard methods to replace hypergraphs by traditional graphs and the loss of information it implies.

\subsection{Hypergraphs and associated graphs}
\label{sec:hyp_def}
\paragraph*{Hypergraphs, definitions}
\textit{Undirected hypergraphs}, sometimes referred to as \textit{hypernetworks} or \textit{2-modes networks}, consist of a set of nodes $V$ and a set of \textit{hyperedges} $E$. Extending the concept of graphs where edges link two distinct nodes, hyperedges are defined as non-empty sets of nodes of arbitrary size. For a hyperedge $e=(u_1, \ldots, u_k)$, its cardinality $k$ is referred to as its $\textit{size}$. A hypergraph in which all hyperedges have the same size $k$ is called a \textit{$k-$uniform hypergraph}. Notably, graphs are $2-$uniform hypergraphs. We further define the star of a vertex $v \in V$ as the set of hyperedges that include $v$: $St(v) = \{ e \in E | v \in e \}$.

\begin{figure}[t]
\begin{center}
\subfigure[$H_1$]{\includegraphics[scale=0.16]{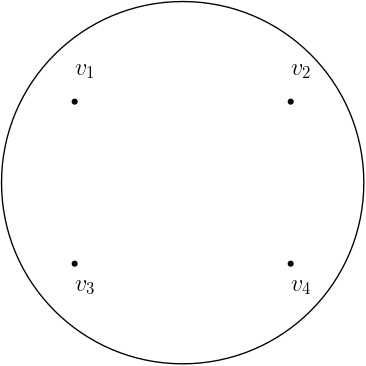}}
\subfigure[$H_2$]{\includegraphics[scale=0.23]{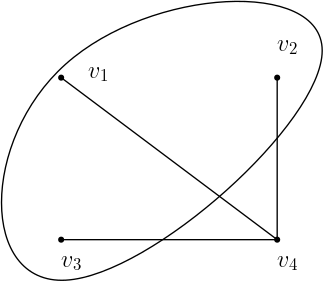}}
\subfigure[$H_3$]{\includegraphics[scale=0.25]{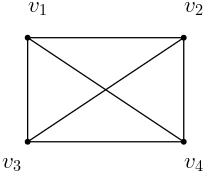}}
\subfigure[$H_4$]{\includegraphics[scale=0.25]{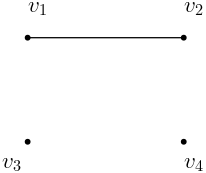}} \\
\subfigure[$\mathbf{C}(H_1)$]{\includegraphics[scale=0.27]{Images/H3.png}} \hspace{0.5cm}
\subfigure[$\mathbf{C}(H_2)$]{\includegraphics[scale=0.27]{Images/H3.png}} \hspace{0.5cm}
\subfigure[$\mathbf{C}(H_3)$]{\includegraphics[scale=0.27]{Images/H3.png}}
\subfigure[$\mathbf{C}(H_4)$]{\includegraphics[scale=0.27]{Images/H4.png}} \\
\subfigure[$\mathbf{L}(H_1)$]{\includegraphics[scale=0.5]{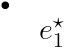}} \hspace{0.7cm}
\subfigure[$\mathbf{L}(H_2)$]{\includegraphics[scale=0.2]{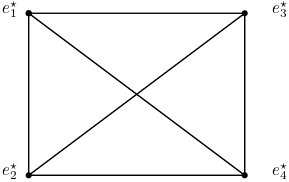}} \hspace{0.5cm}
\subfigure[$\mathbf{L}(H_3)$]{\includegraphics[scale=0.15]{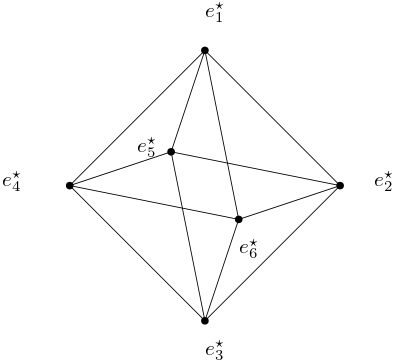}} \hspace{0.5cm}
\subfigure[$\mathbf{L}(H_4)$]{\includegraphics[scale=0.6]{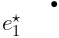}}
\end{center}
\caption{Examples of hypergraphs $H_i$ with their clique $\mathbf{C}(H_i)$ and line $\mathbf{L}(H_i)$ expansions. $H_1$, $H_2$ and $H_3$ share the same clique but have different line expansions.}
\label{ex_HG}
\end{figure}

\paragraph*{Clique expansion}
Given a hypergraph $H$ with node set $V = (v_1, \ldots, v_n)$ and edge set $E = (e_1, \ldots, e_m)$,  its \textit{incidence matrix} $\mathbf{I} \in \{0, 1\}^{n \times m}$ is defined as $\mathbf{I}_{i,j} = 1 $ if $v_i \in e_j$ and 0 otherwise. This representation as a rectangular matrix makes further analysis of hypergraphs much more complicated than that of graphs, which can be described by their adjacency matrix. A commonly used simplification of hypergraphs is the \textit{clique expansion}, where each edge is replaced by a clique. More precisely, the clique expansion $\mathbf{C}(H)$ of the hypergraph $H$ is the graph with node set $V$ and with an edge $(x, y)$ if there exists a hyperedge $e \in E$ such that $x, y \in e$. The clique expansion provides a convenient simplification of hypergraphs by grouping nodes belonging to the same edge. However, there is a certain loss of information since multiple hypergraphs share the same clique expansion, see Figure \ref{ex_HG}. We refer to \cite{hein2013total} for a more thorough analysis of differences between hypergraphs and their clique expansion. Notably, according to \cite{chitra2019random} and \cite{hayashi2020hypergraph}, unless the nodes are assigned an \textit{edge-dependent weighting}, standard notions of random walks and Laplacians on the hypergraphs can be expressed on a reweighted clique expansion. This implies that directly applying Laplacian-based methods, such as clustering, on hypergraphs often incurs the information loss inherent in their clique expansions. We note that the adjacency matrix $A_C$ of the clique graph can be derived from the incidence matrix via $A_C = \mathbf{I} \mathbf{I}^T - D_V$ where $D_v$ is the diagonal node-degree matrix, with entries $d_i = \sum_{j=1}^n \mathbf{I}_{i,j}$.

\paragraph*{Line expansion}
Another graph associated with $H$ is the \textit{line graph} $\mathbf{L}(H)$. Each edge in $H$ now corresponds to a node in $\mathbf{L}(H)$. Thus, $\mathbf{L}(H)$ has node set $(e_1^\star, \ldots, e_m^\star)$ and we have an edge $(e_i^\star, e_j^\star)$ in $\mathbf{L}(H)$ if the corresponding hyperedges intersect in $H$, i.e. $e_i \cap e_j \neq \emptyset$. Similarly to the clique expansion, many hypergraphs share the same line expansion, see Figure \ref{ex_HG}. However, joint knowledge of both the line and clique expansions seems to capture most of the structural information of the hypergraph. Indeed in Figure \ref{ex_HG}, hypergraphs $H_1$, $H_2$ and $H_3$ all share the same clique expansion while having different line expansions (they even have different node sets). Similarly to the clique graph, the adjacency matrix $A_L$ of the line graph can be derived from the incidence matrix via $A_L = \mathbf{I}^T \mathbf{I} - D_e$ where $D_e$ is the diagonal matrix of edge-lengths, with entries $\delta_i = |e_i|$. Let us define the dual hypergraph $H^\star = (E^\star, V^\star )$ as the hypergraph obtained by swapping the edge and node sets, i.e. $E^\star = (e_1^\star, \ldots, e_m^\star)$ and $V^\star = (v_1^\star, \ldots v_n^\star)$ where $v_i^\star = \{ e_j^\star | v_i \in e_j \}$. We can easily observe, see \cite{zhou2022topological}, that the clique expansion is the line graph of the dual hypergraph: $\mathbf{C}(H) = \mathbf{L}(H^\star)$. 

\paragraph*{Gram mates}
We can wonder if the joint knowledge of both the clique graph and the line graph allows to reconstruct the hypergraph. The answer is negative, as there exists pairs of distinct 0-1 matrices $(A, B)$ such that $A A^T = B B^T$ and $ A ^T A = B^T B$. Such matrices are called \textit{Gram mates}, and we refer to \cite{kim2022gram} for a proof of this result and a thorough analysis of families of matrices that generate Gram mates. However, it has been demonstrated in Corollary 1.1.1 of \cite{kirkland2018two} that if we uniformly pick two $m \times n$ 0-1 matrices, the probability that they are Gram mates decays exponentially as $(m,n) \to \infty $. Thus, for practical applications, the information loss from replacing a hypergraph with its clique and line expansions together is negligible. This representation also simplifies hypergraph analysis significantly.

\paragraph*{Clustering}
We consider hypergraphs where nodes are associated with discrete labels $\{1, \ldots, K\}$, representing communities or clusters. Intuitively, such hypergraphs exhibit a community structure, with denser connections between nodes sharing the same label compared to those with different labels. We assume a fully unsupervised setting, where we try to infer the labels (up to permutation) simply via the hypergraph structure. This generalization of the graph partitioning problem has been studied extensively, and we refer to \cite{ccatalyurek2023more} for a survey of different approaches to tackle this problem. A popular method to determine the quality of a partitioning is given by \textit{modularity} maximization, initially defined by \cite{newman2006modularity}. We recall its definition in the graph case. For a graph $G = (V, E)$ and a partitioning $\mathbf{\mathcal{C}} = (C_1, \ldots, C_k)$ of $V$ into $k$ subsets, we define the \textit{modularity function} $Q$ as:

\[ Q (\mathbf{\mathcal{C}}) = \sum_{i=1}^k \frac{e_G(C_i)}{|E|} - \sum_{i=1}^k \left( \frac{vol(C_i)}{vol(V)} \right)^2,
\] 

where $e_G(C_i)$ is the number of edges in the subgraph of $G$ induced by a node set $C_i$, and $vol(A) = \sum_{v \in A} deg(v)$ is the \textit{volume} of any subset $A$ of nodes. Up to renormalization, the modularity function computes the difference between the number of edges uncut by the partitioning and the expected number of edges the same partitioning would yield in the Chung-Lu random graph model introduced in \cite{chung2006complex}.  Maximizing $Q$ over every partitioning yields partitions that minimize cuts within communities, aligning with the underlying structure. There have been several works generalizing modularity for hypergraphs, notably \cite{kumar2020new} and \cite{kaminski2024modularity}, by generalizing the Chung-Lu random model to hypergraphs, and allowing different weights to hyperedges of different sizes. Modularity also serves as a ground truth-free measure of the quality of a given clustering. 

\subsection{Ricci curvature and flows on graphs}
\label{sec:ricci_graph}
This paper addresses the clustering problem using an extension of Ollivier-Ricci curvature to hypergraphs. This method extends prior works on Ollivier-Ricci curvature, initially defined in \cite{ollivier2007ricci} on metric spaces, and extended to the case of graphs in \cite{lin2011ricci}. In \cite{ni2019community}, the Ricci curvature is turned into a discrete flow such that edges between communities are heavily curved. They directly derive a clustering algorithm. We start by recalling the key concepts of their method in the case of graphs.

\paragraph*{Discrete Ricci curvature}

Let $G = (V, E)$ be a weighted graph with a weight function $w: E \to \mathbb{R}_{\geq 0}$. In what follows, the function $w$ is set to be a \textit{dissimilarity measure} between nodes. Let $d$ denote the shortest path distance between nodes induced by weights $w$. We adopt the framework introduced in \cite{ni2019community}: for each node $x$, we define a probability measure $\nu_x^{\alpha, p}$ over $x$ and its neighbors given by:

\begin{equation}
\label{eq:nodes_meas}
\nu_x^{\alpha, p} (y) =
\begin{cases}
\alpha & \text{if } y=x, \\
\frac{1-\alpha} {C} \exp(-d(x,y)^p) & \text{if } y \in \mathcal{N}(x), \\
0 & \text{ otherwise, }
\end{cases}
\end{equation}

where $\mathcal{N}(x)$ denotes the set of neighbors of $x$. Here, $C = \sum_{y \in \mathcal{N}(x)} \exp (-d(x,y)^p)$ is a normalization constant to ensure that $\nu_x^{\alpha, p}$ is a probability measure.  The parameters $\alpha \in [0, 1]$ and $p$ respectively denote the amount of mass retained at $x$ and how much we want to penalize far neighbors. When not required or clear from context, we will drop the superscript and simply write $\nu_x^{\alpha, p} = \nu_x$.

The \textit{discrete Ricci curvature} $\kappa$ is a function defined on the edges of the graph such that for an edge $e=(x, y)$:

\[ \kappa (x, y) = 1 - \frac{W(\nu_x, \nu_y)}{d(x, y)},
\]

where $W$ is the 1-Wasserstein distance between probability measures for the cost function $d$, see \cite{santambrogio2015optimal}. More precisely, a discrete transport plan $T$ between the measures $\nu_x$ and $\nu_y$ is an application $T : V \times V \to [0, 1]$ such that $\sum_{v^\prime \in V} T(u, v^\prime) = \nu_x(u) $ and $\sum_{u^\prime \in V} T(u^\prime, v) = \nu_y(v).$ The 1-Wasserstein distance corresponds to the total cost of moving $\nu_x$ to $\nu_y$ with the optimal transport plan for a cost function $d$, i.e.

\[W(\nu_x, \nu_y) = 	\inf \left\{\sum_{u, v \in V} T(u, v) d(u, v) | T \text{ transport plan between } \nu_x \text{ and } \nu_y \right\}.
\]

\begin{figure}[t]
\begin{center}
\includegraphics[scale=0.15]{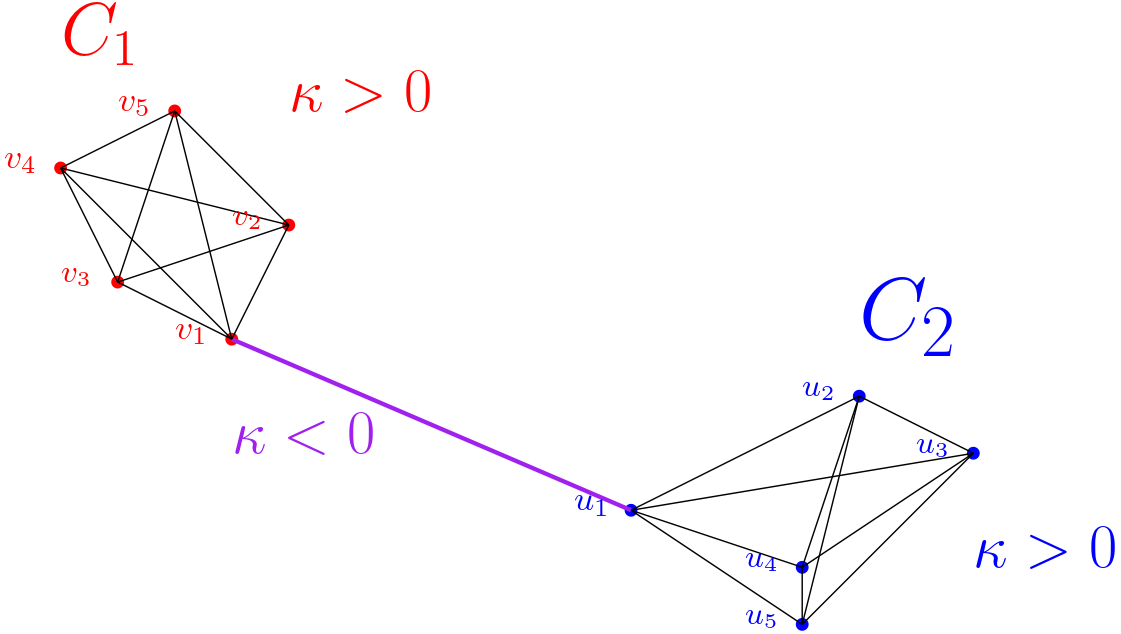}
\end{center}
\caption{Ollivier-Ricci curvature on a graph with a clear two-communities structure.}
\label{fig:ricci_graph_def}
\end{figure}

The sign of $\kappa$ provides information on the density of connections. Indeed, two nodes $x$ and $y$ belonging to the same community will typically share many neighbors, hence transporting $\nu_x$ to $\nu_y$ has a smaller cost than moving $x$ to $y$ and therefore induces a positive curvature. Conversely, nodes belonging to different communities do not share many neighbors, such that moving $\nu_x$ to $\nu_y$ requires traveling through the edge $(x, y)$ bridging the two communities. This in turn corresponds to a negative curvature. We refer to Figure \ref{fig:ricci_graph_def} for a concrete example: when transporting $\nu_{v_i}$ to $\nu_{v_j}$ and $\nu_{u_i}$ to $\nu_{u_j}$, most of the mass remains on the nodes such that measures are transported at a low cost. In the extreme case where $\alpha = 1/5$, all the mass remains and $\kappa (v_i, v_j) = \kappa (u_i, u_j) = 1$. When transporting $\nu_{v_1}$ to $\nu_{u_1}$, all the $v_i$'s must be transported to $u_i$'s through the edge $(u_1, v_1)$ such that the total transport cost is larger than $d(u_1, v_1)$ and therefore $\kappa(u_1, v_1) <0$.

Trimming edges of low curvature typically removes spurious connections between densely connected areas. Consequently, we can obtain a new graph where each connected component corresponds to a different community. We note that Ricci curvature can theoretically be computed for any pair of nodes in the graph. However, for our purposes, we only need to compute it for adjacent nodes.

\paragraph*{Discrete Ricci flow and community detection}
To amplify the impact of heavily negatively curved edges and incorporate more global graph structure, we turn Ricci curvature into a discrete flow that modifies the graph weights. We recall that the edge weights are dissimilarities, and a high weight corresponds to a high distance between nodes. The initial weight between $x$ and $y$ is $w^{(0)}(x, y) = w(x, y)$. We iteratively update the weights with:

\begin{equation}
\label{eq_ric_weights}
w^{(l+1)}(x, y) = \left(1-\kappa^{(l)} (x, y) \right) w^{(l)}(x, y),
\end{equation} 

where $\kappa^{(l)}$ is the Ricci curvature computed with the distance induced by the weights $w^{(l)}$. The updated weights after the $l$-th iteration are referred to as the \textit{Ricci flow}. This flow dynamic stretches edges of low curvature, making it more and more costly to transport measures between communities. The complete clustering algorithm is described by Algorithm \ref{algo_ricci}, and simply trims edges with a high Ricci flow. The choice of the number of flow iterations $N$ is mostly heuristic, we refer to \cite{ni2019community} and the corresponding \texttt{python} library \texttt{GraphRicciCurvature} for some guidelines. The threshold $\tau$ at which to trim the weights is a crucial parameter and can also be taken as the one that maximizes the \textit{modularity function} defined in Section \ref{sec:hyp_def}, which evaluates clustering quality without ground truth labels.

\begin{algorithm}[h!]
\SetAlgoLined
\KwIn{Graph \( G = (V, E) \), Number of iterations \( N \), Threshold \( \tau \)}
\KwOut{Clustering labels \( Y \)}
\For{\( i \gets 1 \) \textbf{to} \( N \)}{
    Update the weights using Equation \ref{eq_ric_weights}\;
}
Trim all edges with a weight larger than \( \tau \) to create a new graph \( \tilde{G} \) ; \\
Compute connected components $C_1 \coprod  \ldots \coprod C_k$ of \( \tilde{G} \) ; \\
\For{\( i \gets 1 \) \textbf{to} \( |V| \)}{
    Set $Y_i = j$ if $v_i \in C_j$;
}
\caption{Ricci flow algorithm.}
\label{algo_ricci}
\end{algorithm}

This approach ensures that inter-community connections are down-weighted while preserving intra-community coherence. This enables a robust and interpretable method for clustering. Indeed, the method relies on direct computations of the Ollivier-Ricci curvature which is strongly associated to an intrinsic community partitioning of the graph, see Figure \ref{fig:ricci_graph_def}. We defer an extended comparison of the interpretability of Ricci-based clustering as opposed to neural network embeddings to Section \ref{sec:expe_real}.

\section{Two complementary notions of Ricci curvature on hypergraphs}
\label{sec:methods}

Extending Ricci curvature to hypergraphs presents two main challenges:
\begin{itemize}
\item Ricci curvature is traditionally defined for pairs of nodes and needs to be extended to hyperedges.
\item Computing transport distances between measures defined on the nodes of a hypergraph typically relies on the clique expansion (see Section \ref{sec:hyp_def} and \cite{hayashi2020hypergraph}). As a consequence, this approach introduces a loss of structural information, as clique expansions fail to capture certain higher-order interactions.
\end{itemize}

The first challenge is straightforward to treat, for instance by taking the average or the maximum of all pairwise curvatures of nodes in a hyperedge (as in \cite{coupette2022ollivier}). We will discuss different possibilities in Section \ref{sec:nodes_transport}. The second challenge constitutes the main contribution of this paper and will further be developed in Section \ref{sec:edges_transport} where we explore a method to transport hyperedges.

Note that we restrict our analysis to the Ollivier-Ricci curvature. Another notion, the \textit{Forman-Ricci curvature} \citep{forman2003bochner}, has been extended to the case of hypergraphs in \cite{leal2021forman}. We refer to \cite{samal2018comparative} for a detailed comparison of these two curvature types in the graph case.

\subsection{Nodes transport on the clique graph}
\label{sec:nodes_transport}
\paragraph*{Clique graph transport}
Let $H$ be a hypergraph with nodes $V = (v_1, \ldots, v_n)$ and edges $E = (e_1, \ldots, e_m)$. To compute the Ricci curvature between two nodes $x$ and $y$, we can still define measures $\nu_x$ and $\nu_y$ on $x$ and $y$ and their neighbors following Equation \ref{eq:nodes_meas}, as the definition of a neighbor is unchanged. We first propose to replace the hypergraph with its clique expansion. Two possibilities are considered to obtain a clique graph from a hypergraph. For the sake of clarity, we assume the hypergraph to be unweighted. Let $A_C = (a_{ij})_{1 \leq i,j \leq n}$ be the adjacency matrix of the clique expansion of $H$.

\begin{enumerate}[(i)]
\item The first option is to consider a simple unweighted clique graph expansion where 
\[
a_{ij} =
\begin{cases}
1 & \text{if there exists } e \in H \text{ such that } (v_i, v_j) \subset e, \\
0 & \text{otherwise}.
\end{cases}
\]

\item The second option is proposed by \cite{zhou2022topological} and relies on a Jaccard index weighting. Recall that for two finite sets $U$ and $V$, the Jaccard index $J(U, V) = \frac{|U \cap V|}{|U \cup V|}$ is a measure of similarity between sets $U$ and $V$. Let $H^\star = (V^\star, E^\star)$ be the dual hypergraph as defined in \ref{sec:hyp_def}. We remark that for two nodes $v_i$ and $v_j$, $a_{ij} = 0$ if and only if $|v^\star_i \cap v^\star_j| = 0 $. The weight of two adjacent nodes in the clique graph is given by $1/J(v^\star_i, v^\star_j)$. The Jaccard index is inverted to provide a dissimilarity weighting.

A discussion on the relative performance of these two weightings can be found in Section \ref{sec:ablation}.

\end{enumerate}

\paragraph*{Aggregating pairwise curvatures}
After computing the Ricci flow $w_\mathcal{N}^{(l)} (e)$ for every pair of adjacent nodes in the clique graph of $H$, the Ricci weight of a given hyperedge $e \in E$ is obtained by aggregating all pairwise flows between nodes of $e$:
\[
w_\mathcal{N}^{(l)} (e) = \underset{v_i, v_j \in e} {Agg} \left(w^{(l)} (v_i, v_j)\right),
\]

where $Agg$ is an aggregation function. Experimentally, we found that aggregating using the \textit{maximum} of all pairwise curvatures systematically yielded the best clustering performance.

The clustering pipeline given by Algorithm \ref{algo_ricci} then operates similarly by trimming all hyperedges with a Ricci flow larger than some threshold. The optimal threshold can be found by either maximizing the modularity of the clique graph, or a corresponding notion of \textit{hypergraph modularity}, see \cite{kaminski2024modularity}.

\subsection{Edges transport on the line graph}
\label{sec:edges_transport}

As discussed in \ref{sec:hyp_def}, replacing a hypergraph by its clique graph induces a substantial loss of information. To address this, we propose a novel notion of Ricci curvature specific to hypergraphs. Instead of transporting measures defined on the nodes of a hypergraph, we transport measures defined on its edges. The motivation is that two nodes from different communities should be contained in sets of hyperedges that typically do not intersect much.

\paragraph*{Edge-Ricci curvature}

The proposed method also computes pairwise curvature and then extends it to hyperedges using an aggregation step similar to the one previously defined. We first describe the Ricci curvature computation for a hypergraph $H = (V, E)$ with initial dissimilarity weights $d: E \to \mathbb{R}_{> 0}$. Let $x$ and $y$ be two adjacent nodes in the hypergraph. We define simple probability measures $\mu_x$ and $\mu_y$ on their stars $St(x)$ and $St(y)$:

\begin{figure}[t]
\begin{center}
\subfigure[Stars of $x$ and $y$]{\includegraphics[scale=0.26]{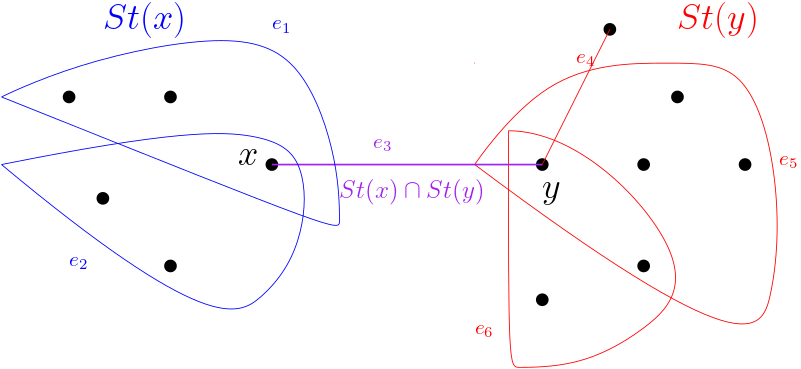}} \hspace*{0.5cm}
\subfigure[Corresponding line graph]{\includegraphics[scale=0.26]{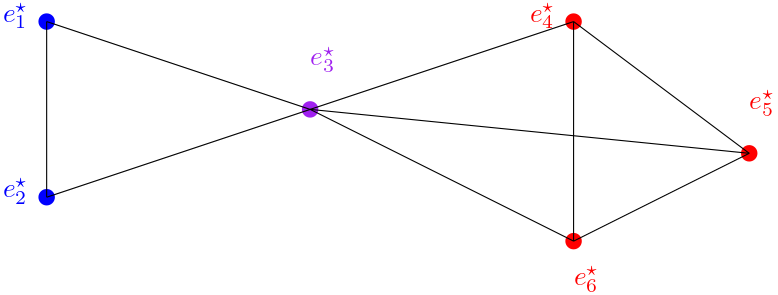}}
\end{center}
\caption{Edge-Ricci transport between nodes $x$ and $y$. A measure on $St(x)$ is transported onto a measure on $St(y)$ via the line graph.}
\label{fig:stars_edge_ricci}
\end{figure}

\begin{equation}
\label{eq:edges_meas}
\mu_x (e) =
\begin{cases}
\frac{d(e)}{C} & \text{if } e \in St(x) \\
0 & \text{ otherwise, }
\end{cases}
\end{equation}


where $C = \sum_{e \in St(x)} d(e) $ is a normalization constant to ensure that $\mu_x$ is a probability measure. The definition of $\mu_y$ is analogous. Similarly to the measures $\nu_x^{\alpha, p}$ defined on the nodes, the measures $\mu_x$ could be adapted to include hyperparameters $\alpha$ and $p$. However, preliminary experiments on a wide variety of settings suggested that it had close to no effect on the resulting curvature.

The distance $d_H$ between two edges $e$ and $f$ is defined using the line graph: $d_H(e, f) = d_{\mathbf{L}(G)} (e^\star, f^\star)$ where $\mathbf{L}(G)$ is the line graph of $H$, see Figure \ref{fig:stars_edge_ricci}. Note that similarly to the clique graph, there are several ways the line graph can be weighted (for instance using Jaccard weights between hyperedges). This aspect will be discussed in more details in Section \ref{sec:ablation}. Similarly to the graph case, we can then define an optimal transport plan and a corresponding 1-Wasserstein distance $W$ using the distance $d_H$ between the two measures $\mu_x$ and $\mu_y$. The \textit{edge-Ricci curvature} $\kappa_{\mathcal{E}}$ between two adjacent nodes $x$ and $y$ is therefore defined as

\[\kappa_{\mathcal{E}}(x, y) = 1-\frac{W(\mu_x, \mu_y)}{\underset{e \in St(x) \cap St(y)}{\max} d(e) }.
\] 

Pairs of nodes in different communities tend to have very different stars, which implies in turn a low curvature. Conversely, if the two nodes $x$ and $y$ belong to the same community, the measures $\mu_x$ and $\mu_y$ can be transported onto each other with a low cost, resulting in a curvature close to 1. We remark that even in the case of a 2-uniform hypergraph, this definition is not equivalent to the usual curvature on graphs from Section \ref{sec:ricci_graph}. In addition, we lose the interpretability observed in Figure \ref{fig:ricci_graph_def} concerning the sign of the curvature. Note that although it is inspired by the notion of curvature developed by \cite{ollivier2007ricci}, edge-Ricci curvature cannot be directly connected to Ollivier's original definition of curvature in metric spaces. As such, deriving theoretical results for this curvature seems currently out of reach. 

As edge-Ricci curvature uses both the clique and the line expansions, it allows to be more sensitive to the hypergraph information than node-Ricci curvature which erases information by considering the clique. For example, we can observe that pairs of nodes in Figure \ref{ex_HG} mastrenghtsy have different edge-Ricci curvature while every pair has the same node-Ricci curvature. However, a general theoretical analysis of the differences between node-Ricci and edge-Ricci curvatures remains out of reach. In Section \ref{sec:theory}, we investigate these differences on a toy example and illustrate the strengths of each notion.

\paragraph*{Edge-Ricci flow}

Similarly to the graph case, the curvature is then turned into a discrete flow. We start with an initial edge weighting $w_{\mathcal{E}}^{(0)} = d$. At the $l$-th iteration, the weights are updated according to the following three steps:

\begin{enumerate}[(i)]
\item Compute the Ricci curvature $\kappa_{\mathcal{E}}^{(l)}$ for every pair of adjacent nodes in the weighted hypergraph.
\item For every hyperedge $e$, aggregate all pairwise curvatures with an aggregation function $Agg$ to define the curvature of the hyperedge, similarly to Section \ref{sec:nodes_transport}:

\[ \kappa_{\mathcal{E}}^{(l)} (e) = \underset{v_i, v_j \in e} {Agg} \left(\kappa_{\mathcal{E}}^{(l)} (v_i, v_j)\right).
\]

\item For every hyperedge $e$, update the edge weights using the Ricci flow dynamic:

\[ w_{\mathcal{E}}^{(l+1)}(e) = \left(1-\kappa_{\mathcal{E}}^{(l)} (e) \right) w_{\mathcal{E}}^{(l)}(e).
\]

\end{enumerate}

Similarly to the graph case this procedure is iterated $N$ times, and hyperedges with a weight larger than a given threshold are trimmed. The resulting connected components form the assigned node communities. In traditional Ricci flow with $p=0$, negatively curved edges gradually gain weight, increasing the cost of moving probability measures through them. Here, we take a dual perspective: edges with low curvature are assigned increasing mass, making them progressively more expensive to transport. The line graph, however, remains unchanged, such that distances between edges are unaffected by the flow process. We refer to \cite{tian2025curvature} for another example of method using the line graph for community detection.

\subsection{A toy clustering example}
\label{sec:theory}

\begin{figure}[t]
\begin{center}
\includegraphics[scale=0.30]{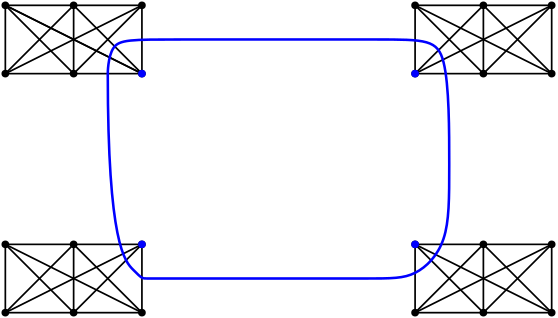}
\end{center}
\caption{Example of hypergraph $H(a,b)$ with $a=6$ and $b=4$. The gateway nodes are represented in blue.}
\label{fig:ex_theo}
\end{figure}

We start by comparing the node-Ricci flow from Section \ref{sec:nodes_transport} with the edge-Ricci flow from Section \ref{sec:edges_transport} in a synthetic example where transport maps and curvatures can be computed explicitly. We adopt a very similar setting to the one of \cite{ni2019community} where we consider a family of hypergraphs $H(a, b)$ where $a, b \geq 3$ are integers. Consider $b$ complete graphs $C_1, \ldots, C_{b}$ on $a$ distinct vertices. For each complete graph $C_i$, take a particular vertex $v_i$ (called gateway vertex) and consider the hyperedge of size $b$, $(v_1, \ldots v_{b})$ connecting all gateway vertices across the communities. This hypergraph exhibits a clear community structure where each $C_i$ forms a distinct community, and there is a single hyperedge connecting all the communities together. Figure \ref{fig:ex_theo} provides a visual representation. 

The case of the node-Ricci flow for an unweighted clique expansion has been treated explicitly in \cite{ni2019community}, for measures defined by Equation \ref{eq:nodes_meas} with $\alpha = 0$ and $p=0$. They observe that by symmetry, in the clique expansion, the Ricci flow at the $l-$th iteration can only take three possible values:
$w_{\mathcal{N}, 1}^{(l)}$ for edges between two gateway nodes, $w_{\mathcal{N}, 2}^{(l)}$ for edges between a gateway node and another node from its community, and $w_{\mathcal{N}, 3}^{(l)}$ for edges between two non-gateway nodes from the same community. They further demonstrate that if $a > b$, $w_{\mathcal{N}, 3}^{(l)} = \left(\frac{1}{a}\right)^l \underset{l \to \infty}{\to} 0$ and that there exists $\lambda > \frac{1}{a}$ and constants $c_1 > c_2$ such that $w_{\mathcal{N}, 1}^{(l)} = c_1 \lambda ^l + o(\lambda^l)$ and $w_{\mathcal{N}, 2}^{(l)} = c_2 \lambda ^l + o(\lambda^l)$ as $l \to \infty$. This implies that the Ricci flow-based trimming algorithm properly identifies the communities $C_1, \ldots, C_{b}$. 

We derive a similar result for specific measures $\mu$. Equation \ref{eq:edges_meas} is slightly modified such that the transported measures are now defined, for two nodes $x$ and $y$ as:

\begin{equation}
\label{eq:edges_meas_alt}
\tilde{\mu}_x (e) =
\begin{cases}
0 & \text{if } x, y \in e \\
\frac{d(e)}{C} & \text{if } e \in St(x) \text{ and } e \notin St(y) \\
0 & \text{ otherwise, }
\end{cases}
\end{equation}

where $C$ is a normalization constant. In practice, this choice of measure yields results extremely similar to the ones defined by Equation \ref{eq:edges_meas}. It simply discards edges shared between $x$ and $y$, which would typically not play a big role in deriving the transport plan. This choice simply ensures a much simpler expression for the Ricci flow to facilitate the delivery of the following proposition: 

\begin{proposition}
\label{prop_toy}
We keep the same notation as above and consider the edge-Ricci flow with measures $\tilde{\mu}$ defined by Equation \ref{eq:edges_meas_alt} and aggregation $Agg$ with the maximum function. At the $l$-th iteration, the Ricci flow can only take three possible values: $w_{\mathcal{N}, 1}^{(l)}$ for the edge of size $b$ connecting all gateway nodes, $w_{\mathcal{N}, 2}^{(l)}$ for binary edges between a gateway node and another node from its community, and $w_{\mathcal{N}, 3}^{(l)}$ for binary edges between two non-gateway nodes from the same community. Under these circumstances, we have that for all $l > 0$:

\[
\begin{cases}
w_{\mathcal{N}, 1}^{(l)} = 2, \\
w_{\mathcal{N}, 2}^{(l)} = \frac{4 + (a-2)w_{\mathcal{N}, 2}^{(l-1)}}{2 + (a-2) w_{\mathcal{N}, 2}^{(l-1)}}  \in [1, 2], \\
w_{\mathcal{N}, 3}^{(l)} = 1.
\end{cases}
\]

\end{proposition}

The proof is deferred to Appendix \ref{sec:appendix_proof}. Proposition \ref{prop_toy} implies that the edge-Ricci flow can also identify the community structure with an edge cut-off parameter between 1 and 2. We notice that in this case, the explicit expression of the Ricci flow, although much simpler than in the node-Ricci case, does not have an exponential decay to 0 for the intra-community edges. This implies that the cut-off parameter might be more delicate to tune in practice. However, note that Proposition \ref{prop_toy} holds for any values of $a$ and $b$, contrary to the result of \cite{ni2019community} which only holds for $a > b$. This implies that in practice, edge-Ricci allows for better detection of many small communities. 

\section{Experiments}
\label{sec:expe}

To evaluate the practical effectiveness of Ricci flow for hypergraph clustering, we conducted a range of experiments on synthetic and real data. We opted for a Jaccard type weighting for the clique graph and the line graph. Curvatures of hyperedges are computed by aggregating curvatures of pairs of nodes using the maximum function. For node-Ricci flow, we have used the python library \texttt{GraphRicciCurvature} with parameters $p=1$ and $\alpha = 0.5$ for the measures $(\nu_x^{\alpha, p})_{x \in V}$ defined on the nodes. We refer to Section \ref{sec:ablation} for a discussion about the choice of default values for the hyperparameters. The clustering accuracy is measured using the \textit{Normalized Mutual Information score} (NMI). The implementations of edge-Ricci and node-Ricci flows for hypergraphs are available at \url{https://github.com/AnonRicciHG/Ricci_curv_HG} and are directly adapted from the \texttt{GraphRicciCurvature} library.

\subsection{Hypergraphs stochastic block models}
\label{sec:synthetic}
We first evaluate our method on a synthetic set-up to highlight differences between edge-Ricci and node-Ricci flows. We consider a hypergraph stochastic block model inspired by \cite{kim2018stochastic}. We consider a hypergraph with $n$ nodes and $k$ communities. We give ourselves two size parameters $s_{in}$ and $s_{out}$. We generate $N_{in}$ hyperedges of size $s_{in}$ by uniformly sampling $s_{in}$ nodes of the same community for each community. We generate $N_{out}$ hyperedges of size $s_{out}$ by first sampling one node from each community at random, and drawing the $s_{out}-2$ remaining nodes uniformly from $N$. In the graph case where $s_{in} = s_{out} = 2$, this is a form of stochastic block model, see \cite{abbe2018community}. This is a simple model where intra-community edges can be interpreted as signal and inter-community edges as noise. In the hypergraph case, as soon as $s_{in} \geq 3$, inter-community edges are also regrouping nodes of the same community together and cannot simply be interpreted as pure noise.

\begin{figure}[hbtp]
\begin{center}
\subfigure[$s_{in} = 2, s_{out} = 2$]{\includegraphics[scale=0.22]{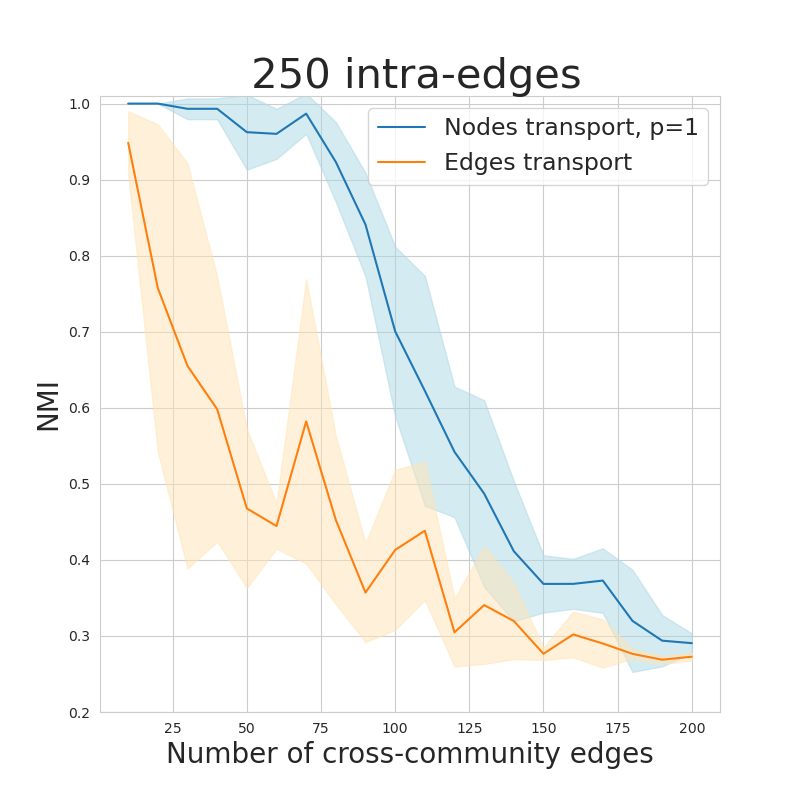}}
\subfigure[$s_{in} = 2, s_{out} = 4$]{\includegraphics[scale=0.22]{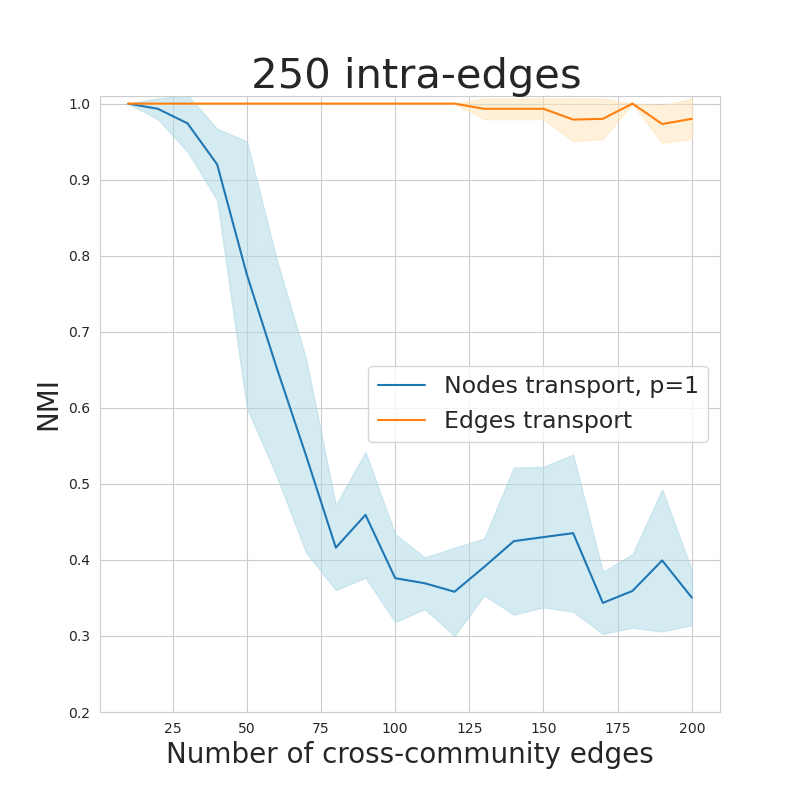}}
\subfigure[$s_{in} = 2, s_{out} = 6$]{\includegraphics[scale=0.22]{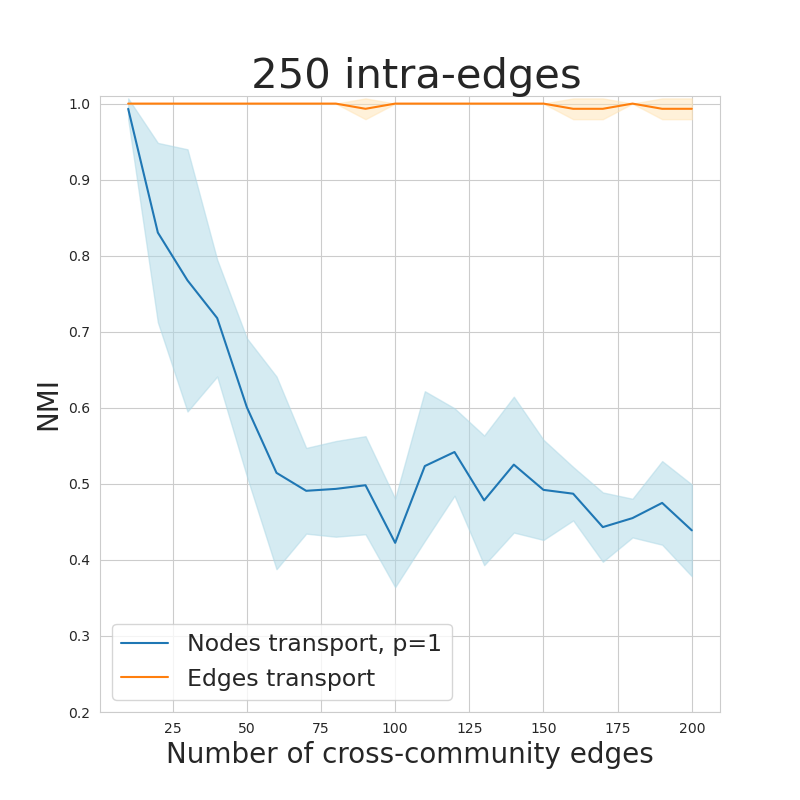}}
 \\
\subfigure[$s_{in} = 3, s_{out} = 2$]{\includegraphics[scale=0.22]{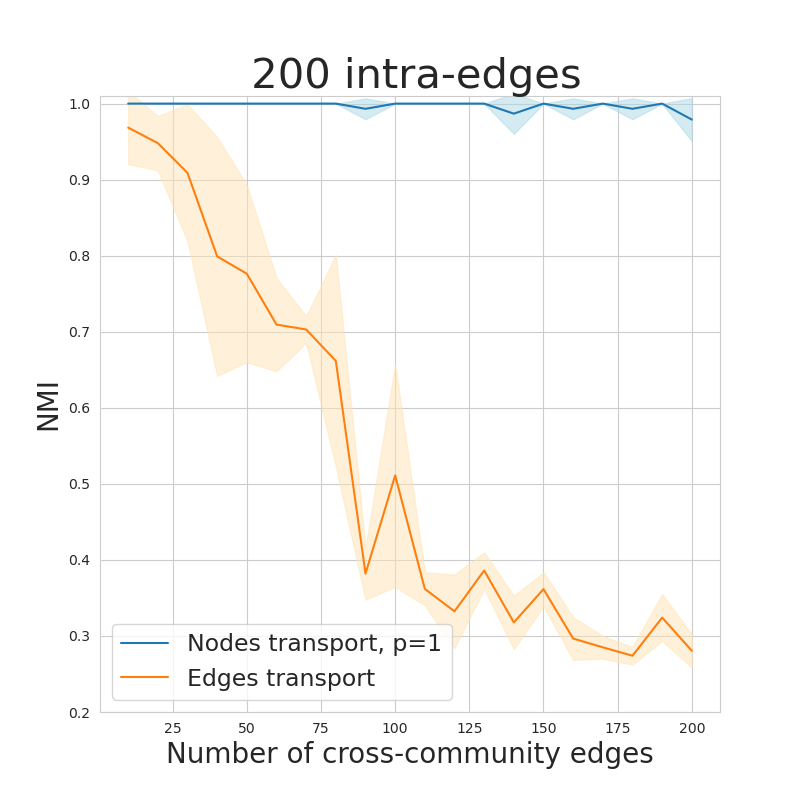}}
\subfigure[$s_{in} = 3, s_{out} = 4$]{\includegraphics[scale=0.22]{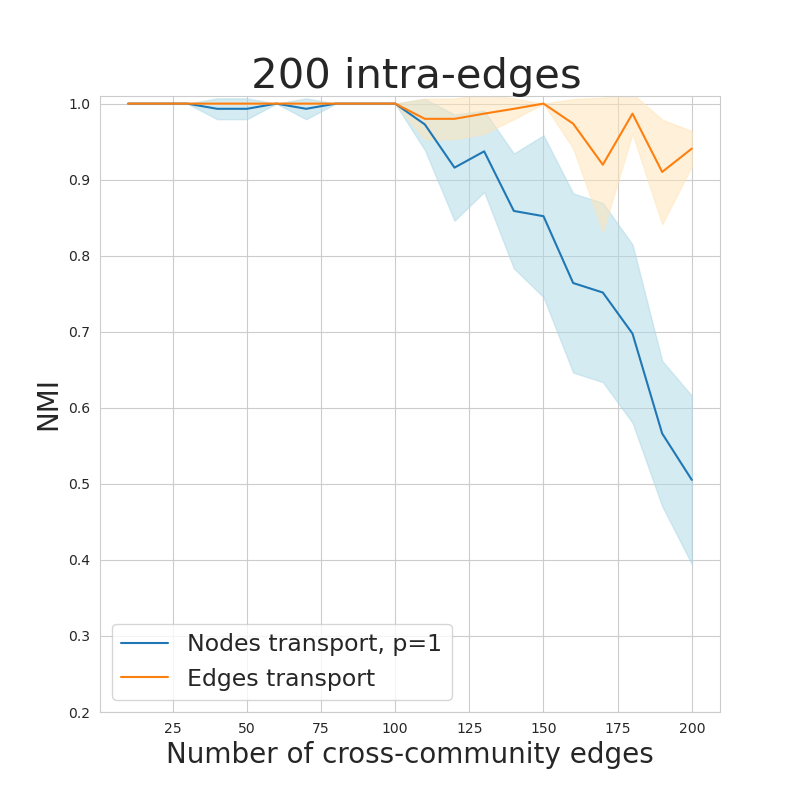}}
\subfigure[$s_{in} = 3, s_{out} = 6$]{\includegraphics[scale=0.22]{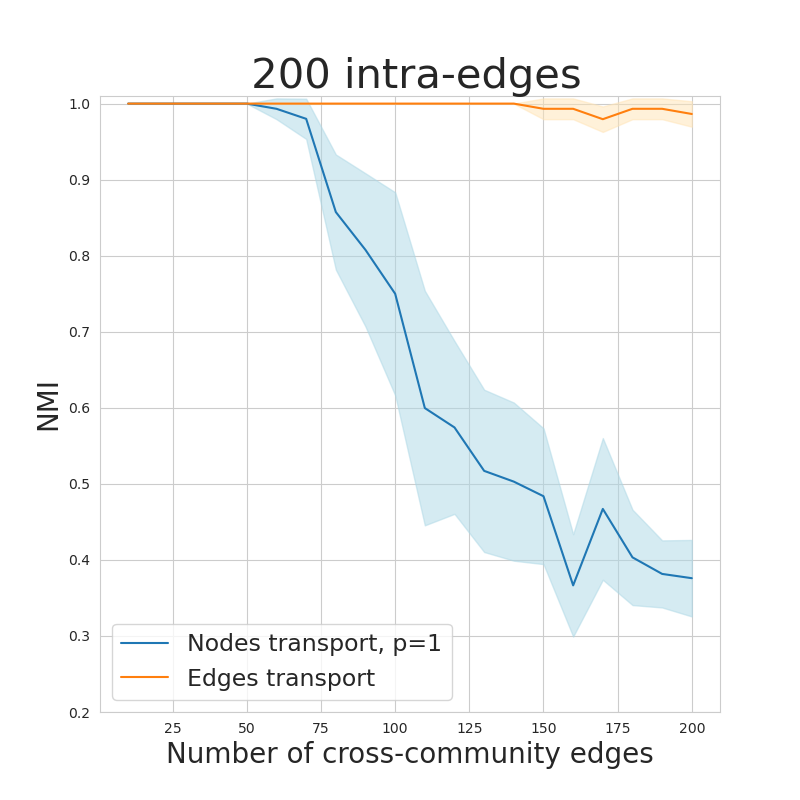}} \\
\subfigure[$s_{in} = 4, s_{out} = 2$]{\includegraphics[scale=0.22]{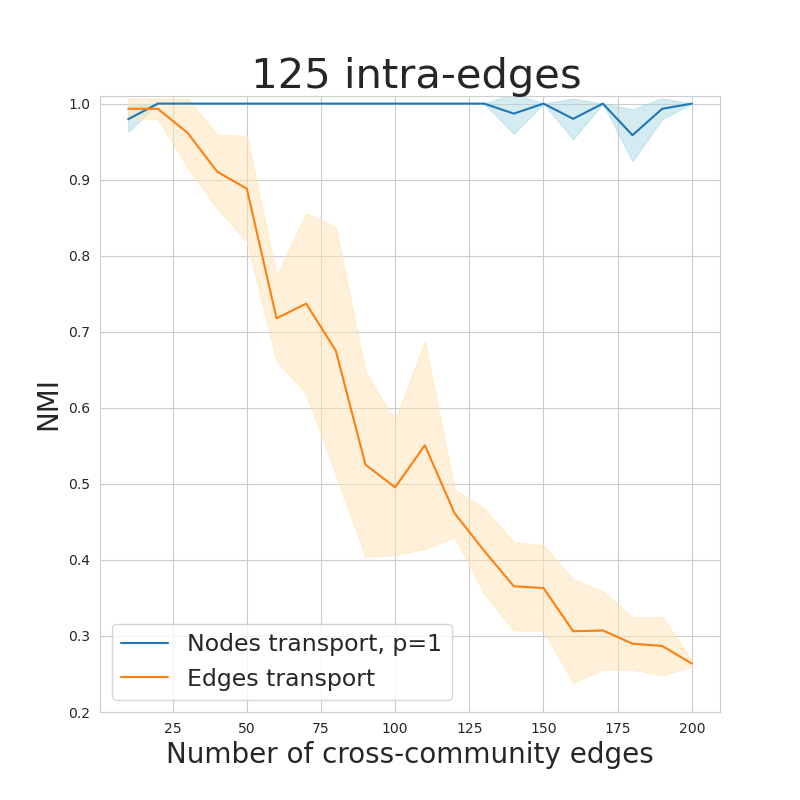}}
\subfigure[$s_{in} = 4, s_{out} = 4$]{\includegraphics[scale=0.22]{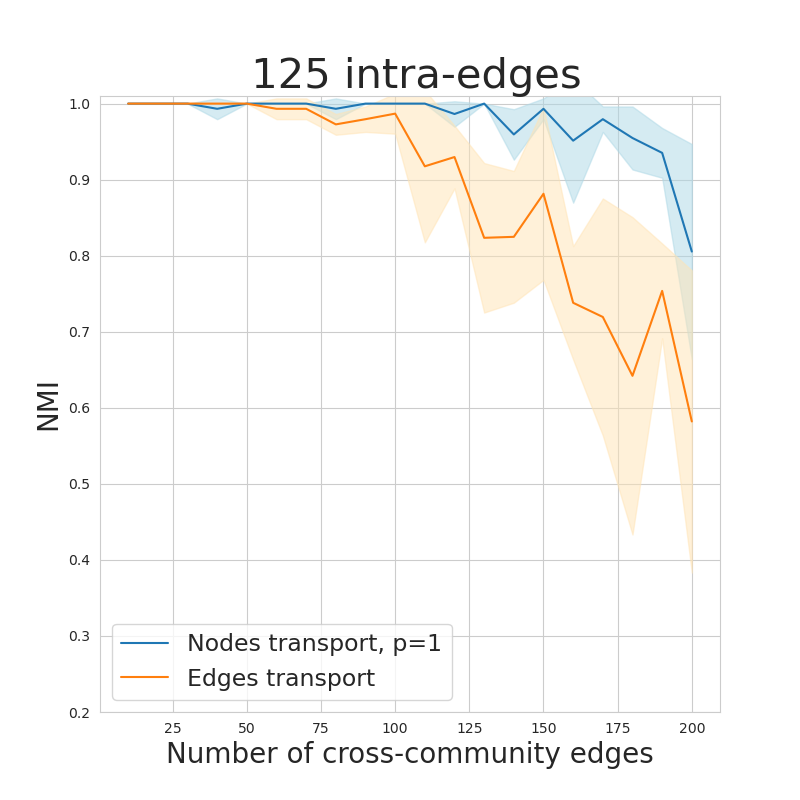}}
\subfigure[$s_{in} = 4, s_{out} = 6$]{\includegraphics[scale=0.22]{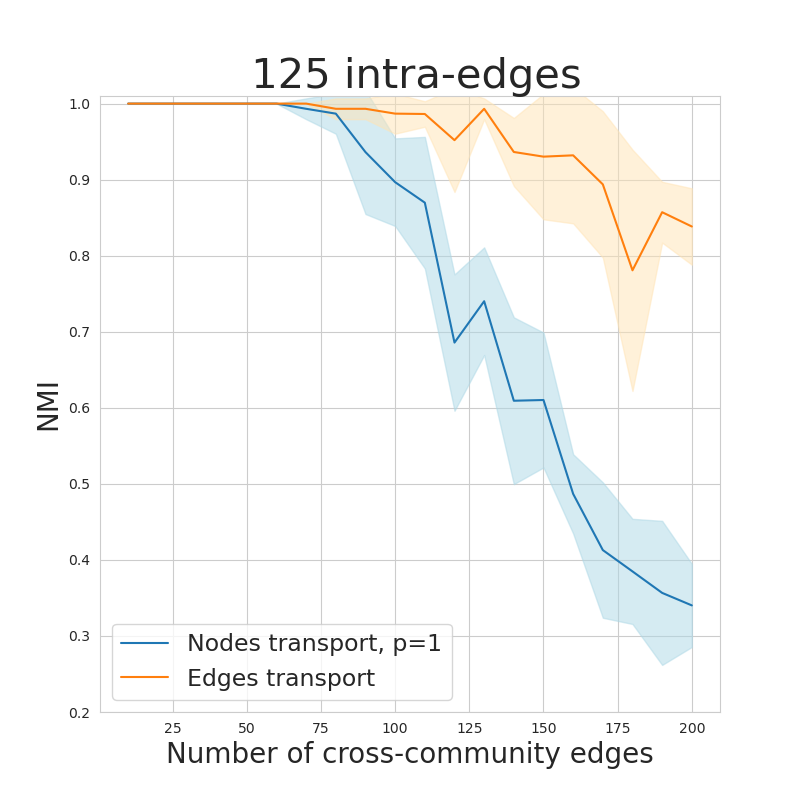}} \\
\subfigure[$s_{in} = 5, s_{out} = 2$]{\includegraphics[scale=0.22]{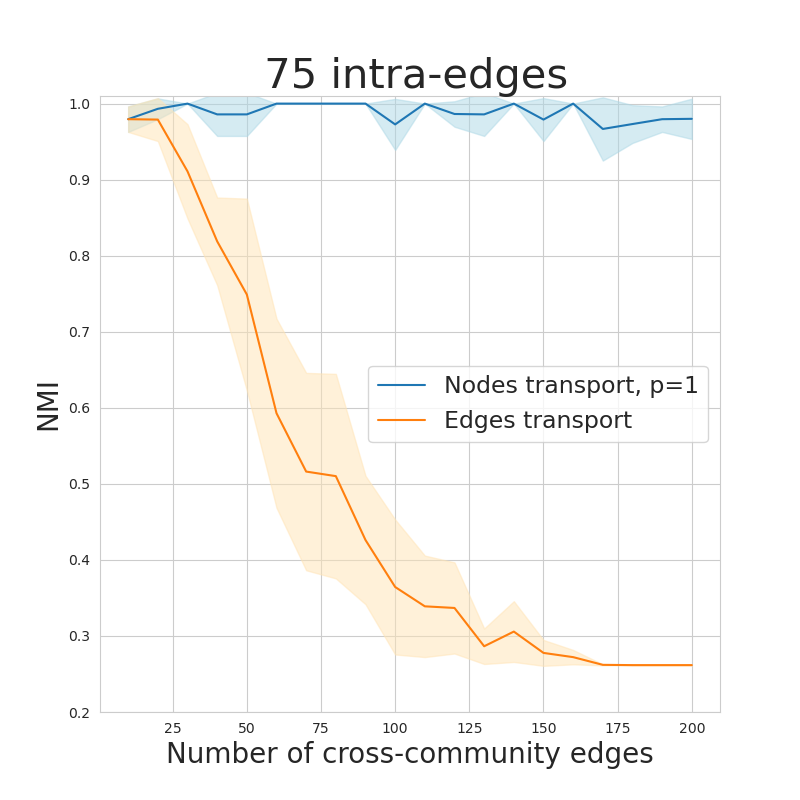}}
\subfigure[$s_{in} = 5, s_{out} = 4$]{\includegraphics[scale=0.22]{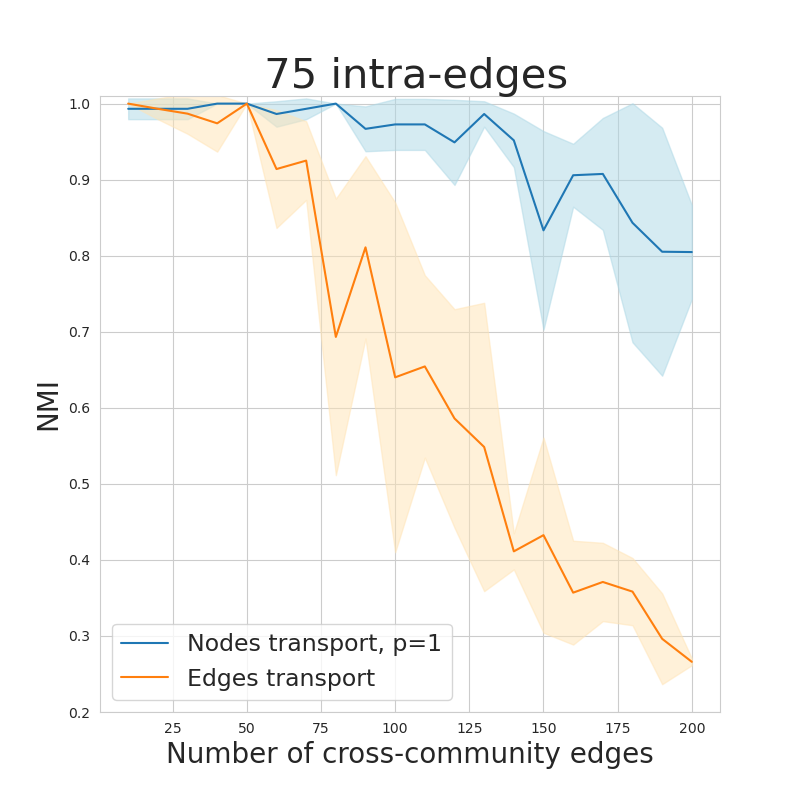}}
\subfigure[$s_{in} = 5, s_{out} = 6$]{\includegraphics[scale=0.22]{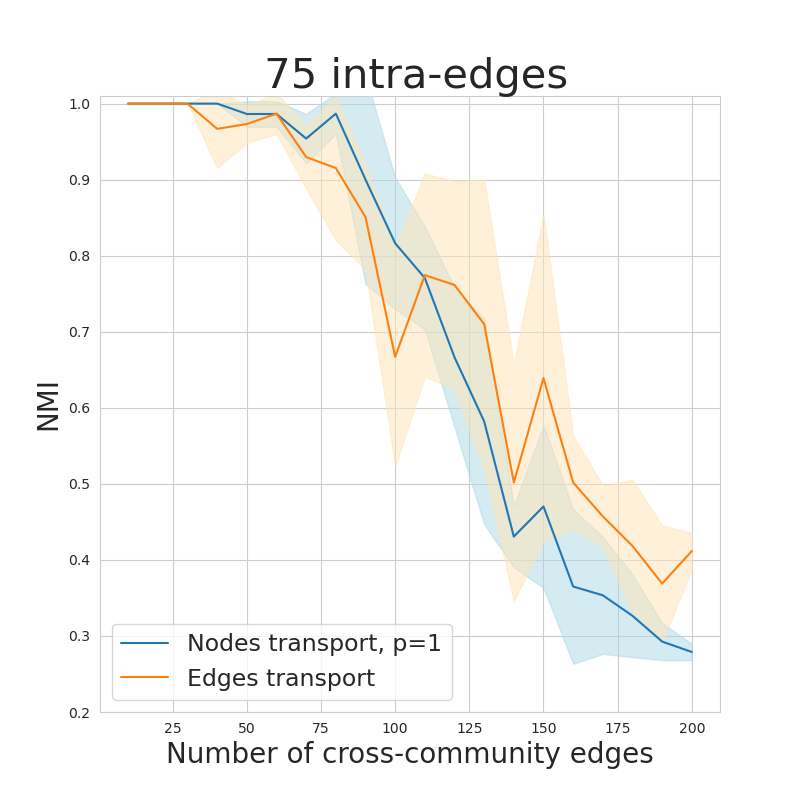}} \\
\end{center}
\caption{Hypergraph stochastic block model reconstruction using hypergraph notions of Ricci flow.}
\label{fig:synthetic}
\end{figure}

We try to reconstruct the communities using edge-Ricci and node-Ricci flows clustering algorithms. We consider randomly generated hypergraphs of $n=100$ nodes, $k=2$ equal-sized communities and compare the performances of the two methods for various values of $s_{in}$ and $s_{out}$. We fix the value of $N_{in}$ and measure the evolution of the clustering accuracy for a growing number $N_{out}$ of "noisy" hyperedges. We compute $N=20$ flow iterations for each method and take the thresholding parameter $\tau^\star$ that maximizes the NMI. We average over 5 random hypergraph generations and report the corresponding NMI in Figure \ref{fig:synthetic}. We remark that node-Ricci flow seems to perform comparatively better than edges-flow when the size of intra-community edges $s_{in}$ is large and the size of inter-community edges $s_{out}$ is small and conversely. These differences can be simply understood heuristically: when contaminating a hypergraph by adding a hyperedge of size $s$, the clique graph will have up to $s(s-1)/2$ new edges while the line graph will only have a single new additional node. This accounts for an improved robustness of edges-flow to large hyperedges contamination, as it mostly relies on shortest-path computations on the line-graph.

\subsection{Real data}
\label{sec:expe_real}
We further compare the clustering performance of nodes and edges-flow with state-of-the-art graph and hypergraph clustering methods. 
\paragraph*{Methodology and data}
We evaluate our method on the datasets presented in Table \ref{tab:data_stats}. A detailed description of the datasets can be found in Section \ref{sec:appendix_datasets}. Data and code to reproduce the experiments are available at \url{https://github.com/AnonRicciHG/Ricci_curv_HG}. We adopt the same benchmark study as in \cite{lee2023m}. For each dataset, we iterate Ricci flows $N=20$ times. The edge-cutting threshold $\tau$ is taken as the one that maximizes the hypergraph modularity, computed with the \texttt{hypernetx} python package. 

\paragraph*{Quantitative results}

We report the NMI of our method in Table \ref{tab:results}, where we compare ourselves with graph and hypergraph partitioning state-of-the-art methods. Node2vec \citep{grover2016node2vec}, DGI \citep{velivckovic2018deep} and GRACE \citep{zhu2020deep} embed the hypergraph's clique expansion in a Euclidean space using various neural networks architectures. $S^2$-HHGR from \cite{zhang2021double} and TriCL from \cite{lee2023m} directly embed the hypergraphs using a neural network architecture. Communities are then detected using a $k$-means algorithm. Note that for TriCL and $S^2$-HHGR, additional node-feature information is used to improve the prediction, while Ricci flow methods rely only on the hypergraph structural information. All scores for these methods are reported from \cite{lee2023m}. We also report the score obtained by the modularity maximization algorithm from \cite{kaminski2021community}. OOM indicates an out-of-memory error on a 24GB GPU (reported from \cite{lee2023m}). OOT indicates that the results could not be obtained within 72 hours on a single 8-core laptop with a Intel(R) Core(TM) i5-8300H CPU @ 2.30GHz processor unit.

\begin{table}[]
\begin{center}
\scalebox{0.9}{
\begin{tabular}{ |c|c|c|c|c|c|c|c| } 

 \hline
 Dataset & Cora-C & Cora-A & Citeseer & Pubmed & Zoo & Mushroom & NTU2012 \\ 
 \hline
 \hline
Node2Vec & 39.1 & 16.0 & 24.5 & 23.1 & 11.5 & 1.6 & 78.3 \\ 
\hline
DGI & $\mathbf{54.8}$ & 45.2 & 40.1 & 30.4 & 13.0 & OOM & 79.6 \\ 
\hline
GRACE & 44.4 & 37.9 & 33.3 & 16.7 & 7.3 & OOM & 74.6 \\ 
\hline
$S^2$-HHGR & 51.0 & 45.4 & 41.1 & 27.7 & 90.9 & 18.6 & 82.7 \\ 
\hline
TriCL & 54.5 & $\mathbf{49.8}$ & $\mathbf{44.1}$ & $\mathbf{30.0}$ & 91.2 & 3.8 & $\mathbf{83.2}$ \\ 
\hline
Modularity & 45.0 & 33.4 & 33.8 & 25.0 & 77.7 & $\mathbf{43.4}$ & 74.5 \\ 
\hline
N-Ricci & 45.8 & 39.4 & 38.8 & 27.8 & 96.2 & OOT & 76.3 \\ 
\hline
E-Ricci & 43.3 & 39.1 & 38.3 & 22.7 & $\mathbf{100.0}$ & 42.3 & 77.6 \\ 
\hline

\end{tabular}}
\caption{NMI clustering accuracy on real datasets.}
\label{tab:results}
\end{center}
\end{table}

Ricci-based clustering methods provide an overall fair accuracy, with a better performance than simple graph vectorization methods, while managing to be as competitive as state-of-the-art neural network-based hypergraph embeddings on a few datasets. In particular, we achieve a perfect clustering on the \texttt{Zoo} dataset using edge-Ricci flow. This dataset stands out for having very few nodes and edges and demonstrates the ability of Ricci-based methods to capture fine structural information on small graphs. We assert that a significant advantage of our method is its high level of explainability, particularly when compared to methods based on neural network embeddings. Specifically, Ricci flow facilitates clustering on graphs with several notable properties:

\begin{itemize}
\item It is grounded in established theoretical works on metric spaces, see \cite{ollivier2007ricci}, and benefits from convergence results when applied on random geometric graphs, as shown in \cite{van2021ollivier}.
\item The method is computable in closed-form on simple examples as demonstrated in \cite{ni2019community} and further elaborated in Section \ref{sec:theory}.
\item It is strongly associated with the intuitive notion of a community on a graph, by assigning a substantial weight to edges across communities, see Figure \ref{fig:ricci_graph_def}.
\item The method relies on explicit and transparent computations and is fully deterministic.
\end{itemize}

However, it is to be noted that when extended to hypergraphs, some of these points fail to be as compelling, notably the theoretical guarantees and the interpretability of the sign of the curvature from Figure \ref{fig:ricci_graph_def}. Nevertheless, our method's explicit computation of a flow process, which draws a strong analogy to its well-established geometric counterpart, offers greater interpretability compared to neural network approaches. The latter often involves many parameters, making their interpretation challenging.

\subsection{Hyperparameter discussion}
\label{sec:ablation}

Both notions of Ricci curvature on hypergraphs depend on a moderate number of hyperparameters. Some of them can significantly influence the final performance. In this section, we review all the hyperparameters, conduct a sensitivity analysis, and provide guidelines for tuning.

\paragraph{The number of flow iterations $N$} 

This parameter is inherited from Ricci flow on graphs. The effect of this parameter has been examined in Section 4.1 of the Supplementary Material of \cite{ni2019community}. The authors experimentally illustrate the convergence of the edge weights as $N$ tend to infinity. Therefore, choosing a large $N$ therefore has no detrimental impact aside from increasing the running time. In the associated code, the authors observe convergence for $N$ between 20 and 50. This guideline still holds in our case, and we did also observe a stabilization of the Ricci flow weights for $N$ larger than 20.

\paragraph{Parameters $\alpha$ and $p$ for $\nu^{\alpha, p}$} 
These parameters are used to define the measures in Equation \ref{eq:nodes_meas}. These parameters are directly inherited from Ricci flow on graphs. The impact of the parameter $p$ has been studied in Section 4.2 of the Supplementary Material of \cite{ni2019community}. We nevertheless decide to review the impact of this parameter in Figure \ref{fig:ablation_p} where we observe that in some cases, there is a drop of performance for $p=0$ while $p=1$ and $p=2$ show a similar performance. Additionally, we decide to investigate the role of the parameter $\alpha$ in some critical cases, see \ref{fig:ablation_alpha}. This study has not been conducted in loc. cit. where the authors advocate for a default choice of $\alpha = 0.5$. Although it is challenging to provide specific guidelines, extreme values of the parameter $\alpha$ can lead to serious instabilities. As a result, taking $\alpha$ close to 0.5 is a safe choice.

\paragraph{Aggregation and graph weightings}
Both methods compute a Ricci-curvature between pairs of adjacent nodes which is in turn aggregated at the hyperedge level, as suggested in \cite{coupette2022ollivier}. We have explored two choices: taking the maximum and the arithmetic average. 

In addition, the transport costs depend on the initial clique graph and line graph weightings discussed in Section \ref{sec:nodes_transport}. We compare between unweighted graphs and Jaccard weightings. We have studied the joint impact of these two parameters and report the results in Figures \ref{fig:ablation_jac_max_nodes} and \ref{fig:ablation_jac_max_edges}. We make the following observation:
\begin{itemize}
\item For node-Ricci flow, the clique graph weighting has little to no impact, while using the maximum of the pairwise curvature for the hyperedge curvature consistently yields better results.
\item For edge-Ricci flow, we observe a very important difference between these parameters. In short, the combination Jaccard-maximum is the most competitive when $s_{in} \leq s_{out}$ and Uniform-average is the most competitive when $s_{in} > s_{out}$. The other two combinations always yield an intermediate accuracy. This implies that an intersection-sensitive line graph weighting has a strong positive impact in the case of large hyperedges between communities. Nevertheless, we observe in Figure \ref{fig:ablation_comparison_NE} that edge-Ricci flow with any combination of parameters is always outperformed by a node-Ricci flow with a maximum aggregation function in the case where $s_{in} > s_{out}$, as already observed in Section \ref{sec:synthetic}. Therefore, we always recommend using the Jaccard-maximum parameter combination for the edge-Ricci flow to fully leverage its potential in the case of large inter-community hyperedges.
\end{itemize}

\paragraph{Joint study of every hyperparameter on a real dataset}
In the previous paragraphs, we considered critical cases where a change in parameters could have a dramatic impact. We now study the robustness of the two types of transport applied to a real dataset. In Tables \ref{tab:ablation_N} and \ref{tab:ablation_E}, we examine the joint effect of every parameter except $\alpha$ and $\tau$ on the dataset \texttt{NTU2012}. We observe that for this real dataset, the choice of weighting has almost no impact on the final accuracy. However, using the maximum function for $Agg$ provides a much better accuracy than the average. Finally, We observe a modest improvement by considering $p=1$ over $p=0$.

\paragraph{Threshold parameter $\tau$}
Both notions of Ricci flow simply operates a community-sensitive reweighting of the edges, equivalent to a hierarchical clustering of the nodes. The cut-off parameter $\tau$ enables access to a particular clustering, making it a crucial parameter when comparing to ground-truth labels. To chose the value of $\tau$, we compare the one that maximizes the hypergraph modularity $\tau_{H}$, the one that maximizes the graph modularity on the clique expansion $\tau_{C}$ and the one giving the clustering with the highest NMI, i.e. closest to the ground-truth one $\tau^\star$. We report these results in Tables \ref{tab:tau_ntu} and \ref{tab:tau_zoo}. We observe that using $\tau_H$ gives results relatively close from the optimal ones using $\tau^\star$ and better accuracy than using $\tau_C$. In particular, on the \texttt{Zoo} dataset, $\tau_H=\tau^\star$ for both notions of transport, while $\tau_C$ gives extremely poor results. This additional experiment aligns with the findings of \cite{kaminski2021community} and \cite{lee2023m}, advocating for using hypergraph-specific tools when available, rather than reducing to the clique expansion.

Finally, note that there is not necessarily a single best value for the parameter $\tau$. Indeed, modifying the optimal $\tau$ can provide clusters coarser or finer than the ground-truth, which can be appealing in some applications. 

\subsection{Computational aspects}
\label{sec:complexity}

\paragraph*{Comparison between the two types of transport}

We observe in Table \ref{tab:results} that node-Ricci flow failed to provide results for the \texttt{Mushroom} dataset in a reasonable time. Even after 72 hours, not a single Ricci-flow iteration was completed, despite this dataset containing a relatively small number of hyperedges (see Table \ref{tab:data_stats}). To better understand this limitation, we further investigate the computational complexity of both methods. 

Computing the Ricci curvature of all edges in a hypergraph $H = (V, E)$ requires solving $|\mathcal{E}|$ optimal transport problems where $\mathcal{E}$ is the edge set of the clique expansion $\mathbf{C}(H)$. In practice, the optimal transport costs between two measures with at most $n$ points are computed using the Sinkhorn algorithm from \cite{sinkhorn1974diagonal}, which provides an $\varepsilon$-approximation at a cost $O(n^2/\varepsilon^2)$ according to \cite{lin2019efficient}. This implies worst-case complexities of order $O \left(|\mathcal{E}| \times \underset{x \in V}{\max} |\mathcal{N}(x)|^2/\varepsilon^2 \right)$ for node-Ricci and $O \left(|\mathcal{E}| \times \underset{x \in V}{\max} |St(x)|^2/\varepsilon^2 \right)$ for edge-Ricci. In turn, large hyperedges have a stronger impact on the node-Ricci complexity.

\begin{figure}[t]
\begin{center}
\includegraphics[scale=0.25]{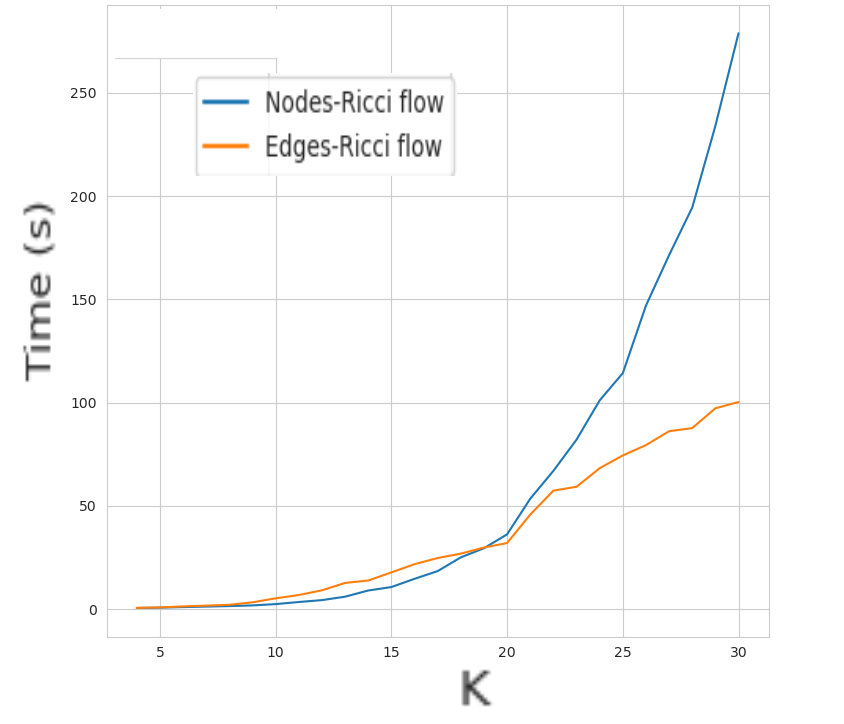}
\end{center}
\caption{Computation times of Ricci curvature for both methods on a $K$-uniform hypergraph with 1000 nodes and 300 edges as a function of $K$.}
\label{fig:time}
\end{figure}

To justify these theoretical claims, we consider a $K$-uniform hypergraph with 1000 nodes and 300 edges of length $K$ drawn uniformly at random. We report the computational times in seconds for a single Ricci-flow interation in Figure \ref{fig:time}. Computations are carried out for unweighted clique and line expansions and are averaged over five different hypergraph initializations. We observe that the computational cost of nodes-Ricci curvature indeed blows up for hypergraphs with large hyperedges due to the inflated neighborhood size $|\mathcal{N}(x)|$. Edge-Ricci flow demonstrates much slower growth in computational time, remaining efficient even for hypergraphs with large hyperedges. Computations are executed on a single 8-core laptop with a Intel(R) Core(TM) i5-8300H CPU @ 2.30GHz processor unit. These computations explain why only edge-Ricci could produce results in a reasonable time on the \texttt{Mushroom} dataset. Although this dataset has relatively few hyperedges, the hyperedges are exceptionally large, inducing severe computational challenges for node-Ricci flow. 

\paragraph*{Comparison on real data}

We now compare the two notions of Ricci flow to the modularity-based clustering. As they involve solving many optimal transport problems, flow-based methods appear to be quite costly in practice. We report computational times for $N=20$ flow iterations in Table \ref{tab:time_real}. For datasets which typically have nodes with large stars (especially citation datasets), the computational performance of edge-Ricci can be disadvantageous in some cases when compared to modularity or node-Ricci. The times from Table \ref{tab:time_real} together with the quantitative results from Table \ref{tab:results} indicate that Ricci-flow based methods are more suited to dealing with rather small hypergraphs to extract fine information on them.

\begin{table}[t]
\begin{center}
\scalebox{0.9}{
\begin{tabular}{ |c|c|c|c|c| } 

 \hline
 Dataset & Cora-C & Pubmed & NTU2012 & Zoo \\ 
 \hline
 \hline
Modularity & 150.1 & 298 & 40.1 & 0.9 \\ 
\hline
N-Ricci & 43 & 7890 & 48.8 & 7.4 \\ 
\hline
E-Ricci & 296 & 31500 & 40.5 & 17.7 \\ 
\hline

\end{tabular}}
\caption{Computational time on real data (in seconds)}
\label{tab:time_real}
\end{center}
\end{table}

\subsection{Wrap-up, comparison of Node-Ricci and Edge-Ricci flows}

Based on the findings from Sections \ref{sec:theory} and \ref{sec:expe}, node-Ricci and edge-Ricci flows are complementary approaches for competitive hypergraph clustering. Indeed, as the first one primarily leverages the clique expansion and the second one the line expansion information, taken together, they capture most of the information of the hypergraph. More precisely, edge-Ricci is preferable to node-Ricci whenever:
\begin{itemize} 
\item There are many small communities.
\item The intra-community edges are typically smaller than inter-community edges.
\item The hypergraph has very large hyperedges, inducing prohibitive computations on the clique graph.
\end{itemize}

When dealing with real data with no structural a priori on the communities, both methods tend to perform comparatively, see Section \ref{sec:expe_real}. However, they can have sensibly different computational costs making one preferable over the other depending on the structure of the hypergraph, in particular the number of hyperedges and their size. We note that alternating between iterations of node-Ricci flow and edge-Ricci flow did not provide any substantial quantitative benefit. 

\section*{Conclusion}

We have developed two methods to extend Ricci flow to hypergraphs. The first method, node-Ricci flow, applies standard Ricci flow on the clique expansion and further aggregates it to hyperedges. The second method, edge-Ricci flow, is the main contribution of this article and constitutes an original approach to transporting edges using the line expansion. Both methods define new weights on the hyperedges, which can then be simply turned into a partitioning algorithm. Each method has its own advantages depending on the specific characteristics of the hypergraph. 

A natural extension of this work would be to consider a co-optimal transport of both nodes and edges, as discussed in \cite{titouan2020co, chowdhury2024hypergraph}. Additionally, the notion of discrete Ricci flow has more practical uses beyond graph partitioning. For instance, it can be used as a method to preprocess hypergraphs by generating new weights that better highlight underlying community structures, as seen in \cite{khodaei2024classification}. Finally, several notions of curvature have been developed for graph, as explored in \cite{iyer2024non, cushing2022bakry}. Further extending these notions to hypergraph data constitutes a potential research direction.

%
%
%
\bibliographystyle{apalike}
\bibliography{biblio_ricci}

\appendix
\section{Proof of Proposition \ref{prop_toy}}
\label{sec:appendix_proof}
\begin{itemize}
\item We start by computing $w_{\mathcal{N}, 3}^{(l)}$. Let $x$ and $y$ be two non-gateway nodes from the same clique $(x, y, u_1, \ldots, u_{a-2})$. We have that $St(x) \setminus (x, y) = \{(x, u_1), \ldots, (x, u_{a-2}) \}$ and $St(y) \setminus (x, y) = \{(y, u_1), \ldots (y, u_{a-2}) \}$. Each $(x, u_i)$ is adjacent to $(y, u_i)$ and at a distance two from the others $(y, u_j)_{j \neq i}$? The optimal way to transport the uniform probability measure on $St(x) \setminus (x, y)$ to the uniform probability measure on $St(y) \setminus (x, y)$ is therefore to transport each $(x, u_i)$ to $(y, u_i)$, which are at a distance 1 from each other. Each edge has weight $1/(a-2)$ and there is $a-2$ of them, hence a total cost of $1$. Hence by induction, since $w_{\mathcal{N}, 3}^{(l)} = 1$, we have that $w_{\mathcal{N}, 3}^{(l)} = 1$ for all $l$.

\item For $w_{\mathcal{N}, 1}^{(l)}$, let $x$ and $y$ be two distinct gateway-nodes, connected by the large hyperedge $E_b$. We have $St(x) \setminus E_b = \{ (x, u_1), \ldots, (x, u_{a-1}) \}$ and $St(y) \setminus E_b = \{ (y, v_1), \ldots, (x, v_{a-1}) \}$. Each $(x, u_i)$ is at distance at least 2 from a $(y, v_i)$. Therefore, an optimal way to transport the uniform probability measure on $St(x) \setminus E_b$ to the uniform probability measure on $St(y) \setminus E_b$ is to transport each $(x, u_i)$ to a corresponding $(y, v_i)$ via $E_b$, such that the distance is exactly 2. This corresponds to a path of length $2$ on the line graph. Each edge has weight $1/(a-1)$ and there is $a-1$ of them, hence a total cost of $2$. As the cost is the same for each pair of nodes in $E_b$, the flow of $E_b$ is equal to that of any pair $(x,y) \in E_b$. We therefore have that $w_{\mathcal{N}, 3}^{(l)} = 2$ for all $l \geq 1$.

\item For $w_{\mathcal{N}, 2}^{(l)}$, let $x$ be a gateway node and $y$ and non-gateway node from the same clique $(x, y, u_1, \ldots, u_{a-2}$. We have $St(x) \setminus (x, y) = \{ E_b, (x, u_1), \ldots, (x, u_{a-2}) \}$ and $St(y) \setminus (x, y) = \{(y, u_1), \ldots, (y, u_{a-2}) \}$. We represent the line graph restricted to $St(x) \cup St(y) \setminus (x,y)$ in Figure \ref{fig:LG_proof}. We write the mass of each edge according to Equation \ref{eq:edges_meas} where we write $d_i = w_{\mathcal{N}, i}^{(l-1)}$ to simplify the notation. We have just demonstrated that $d_1 = 2$ and $d_3 = 1$. From Figure \ref{fig:LG_proof}, it is clear than an optimal transport plan between $\mu_x$ and $\mu_y$ transports the mass on each $(x, u_i)$ to $(y, u_i)$. These edges are adjacent in the line graph. These $(a-2)$ edges therefore occupy a fraction of the total cost of $\frac{(a-2)d_2}{2+(a-2)d_2}$. The mass $\frac{d_2}{2 + (a-2)d_2}$ on $E_b$ is equally transported to each $(y, u_i)$ through the corresponding $(x, u_i)$. Each $(y, u_i)$ is at a distance $2$ from $E_b$, see Figure \ref{fig:LG_proof}. This implies a cost of $\frac{2d_2}{2+(a-2)d_2}$ for transporting the mass at $E_b$. In the end, the total cost of transporting $\mu_x$ to $\mu_y$ is $\frac{4+(a-2)d_2}{2+(a-2)d_2}$. This implies that

\[
w_{\mathcal{N}, 2}^{(l)} = \frac{4 + (a-2)w_{\mathcal{N}, 2}^{(l-1)}}{2 + (a-2) w_{\mathcal{N}, 2}^{(l-1)}}.
\]

We can easily prove by induction on $l$ that for every $l$, $w_{\mathcal{N}, 2}^{(l)} \in [1, 2]$.

\renewcommand\thefigure{S.\arabic{figure}}
\setcounter{figure}{0}

\begin{figure}[t]
\begin{center}
\includegraphics[scale=0.40]{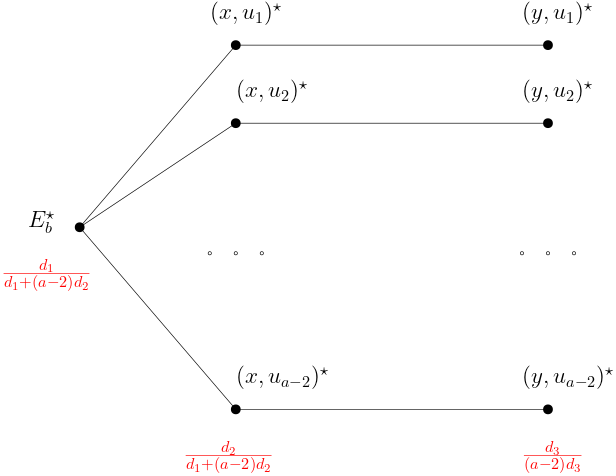}
\end{center}
\caption{Restriction of the line graph to compute the flow between a gateway and a non-gateway node in Proposition \ref{prop_toy}.}
\label{fig:LG_proof}
\end{figure}
\end{itemize}

\section{Description of the datasets}
\label{sec:appendix_datasets}

The experiments from Section \ref{sec:expe_real} have been conducted on these hypergraph datasets:

\begin{itemize}

\item Co-citation datasets \texttt{Cora-C}, \texttt{Citeseer} and \texttt{Pubmed} where nodes represent research papers, edges papers co-citing these papers and the communities correspond to similar research thematic. 
\item The dataset \texttt{Cora-A} has a similar structure but is a co-authorship dataset, where hyperedges represent authors. 
\item In the dataset \texttt{Zoo}, nodes are animals, edges represent mutual traits, and communities are types of animals (mammal, fish...). 
\item In the dataset \texttt{Mushroom}, nodes are mushroom samples, edges represent mutual traits, and there are two communities: edible and poisonous. 
\item In the dataset \texttt{NTU2012}, nodes correspond to 3D objects, edges to proximity using image-processing features, and communities are the type of object. 

\end{itemize}

We refer to Table \ref{tab:data_stats} for a quantitative description of each dataset.

\renewcommand\thetable{S.\arabic{table}}
\setcounter{table}{0}
\begin{table}[h!]
\begin{center}
\scalebox{0.9}{
\begin{tabular}{ |c|c|c|c|c|c|c|c| } 

 \hline
 Dataset & Cora-C & Cora-A & Citeseer & Pubmed & Zoo & Mushroom & NTU2012 \\ 
 \hline
 \hline
$\sharp$ Nodes & 1434 & 2388 & 1458 & 3840 & 101 & 8124 & 2012 \\ 
\hline
$\sharp$ Hyperedges & 1579 & 1072 & 1079 & 7963 & 43 & 298 & 2012 \\ 
\hline
Avg. hyperedge size & 3.0 & 4.3 & 3.2 & 4.4 & 39.9 & 136.3 & 5.0 \\ 
\hline
Avg. node degree & 3.3 & 1.9 & 2.4 & 9.0 & 17.0 & 5.0 & 5.0 \\ 
\hline
$\sharp$ Communities & 7 & 7 & 6 & 3 & 7 & 2 & 67 \\ 
\hline

\end{tabular}}
\caption{Basic statistics of hypergraph data used in the experiments.}
\label{tab:data_stats}
\end{center}
\end{table}

\section{Impact of the hyperparameters}
\label{sec:appendix_ablation}

Unless specified otherwise, we report the NMI in the synthetic set-up of Section \ref{sec:synthetic}. Results are averaged over 5 random hypergraph generation and comparison with the ground-truth is realized for the optimal cut-off parameter $\tau^\star$.

\paragraph{Measure parameter $p$.}

In Figure \ref{fig:ablation_p}, we report the NMI for three values $p=0, 1, 2$ for the nodes transport, compared with the edge transport. We set $Agg$ to be the maximum function and consider a uniform weighting of the clique graph. We can see that in some cases, there is a clear drop of accuracy for $p=0$, while $p=1$ and $p=2$ display similar performances.

\renewcommand\thefigure{S.\arabic{figure}}
\begin{figure}[hbtp]
\begin{center}
\subfigure[$s_{in} = 2, s_{out} = 3$]{\includegraphics[scale=0.19]{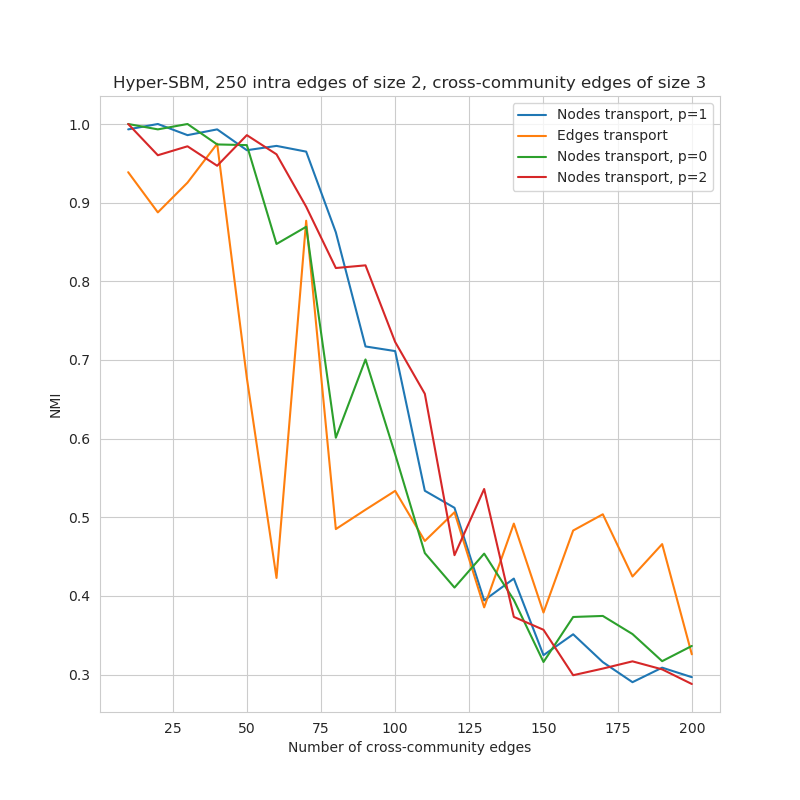}}
\subfigure[$s_{in} = 3, s_{out} = 3$]{\includegraphics[scale=0.19]{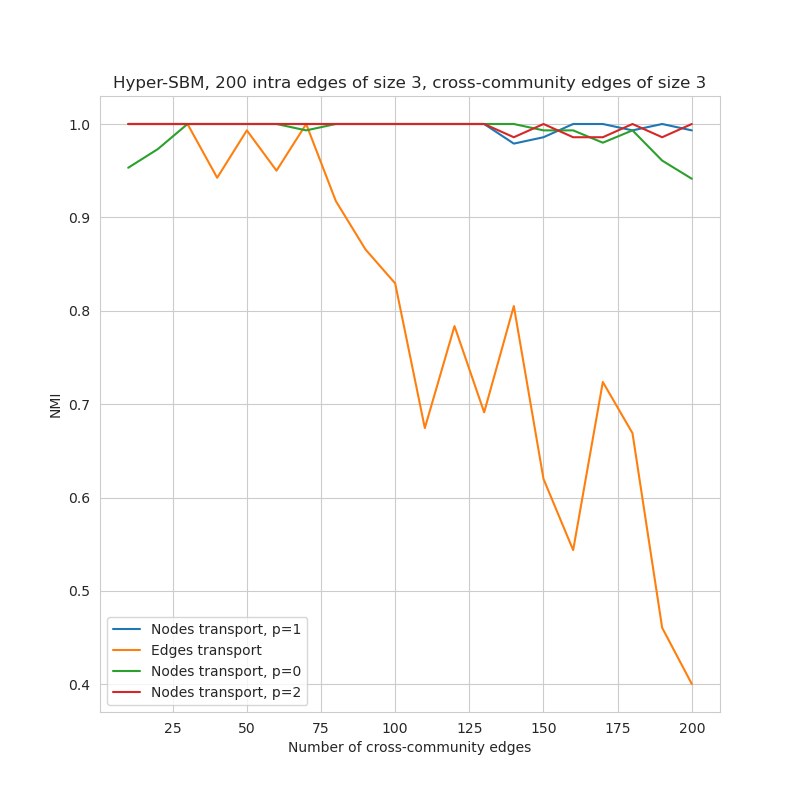}}
\subfigure[$s_{in} = 4, s_{out} = 3$]{\includegraphics[scale=0.19]{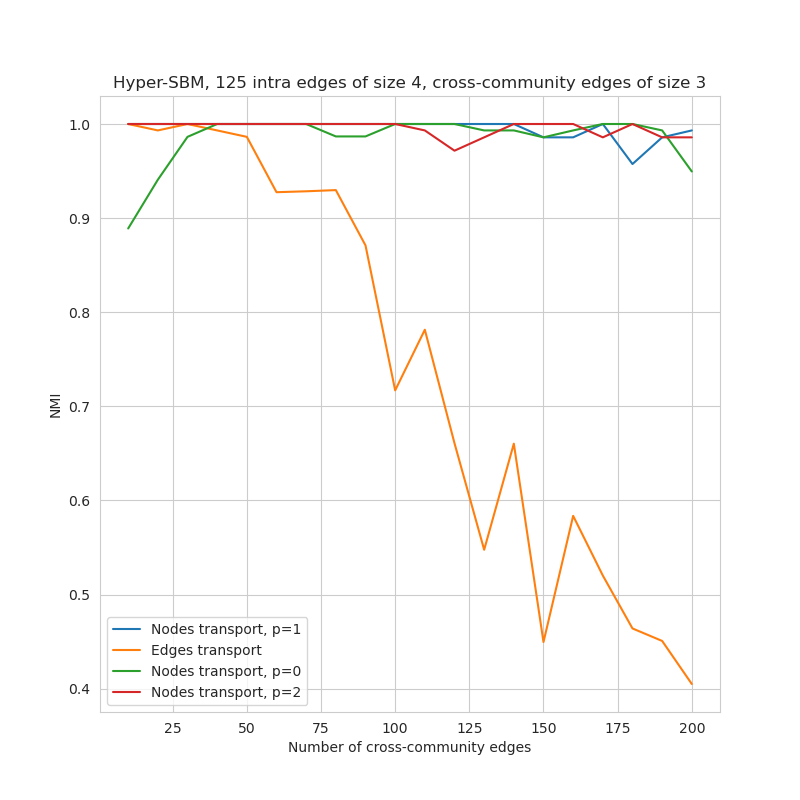}}
 \\
\subfigure[$s_{in} = 2, s_{out} = 5$]{\includegraphics[scale=0.19]{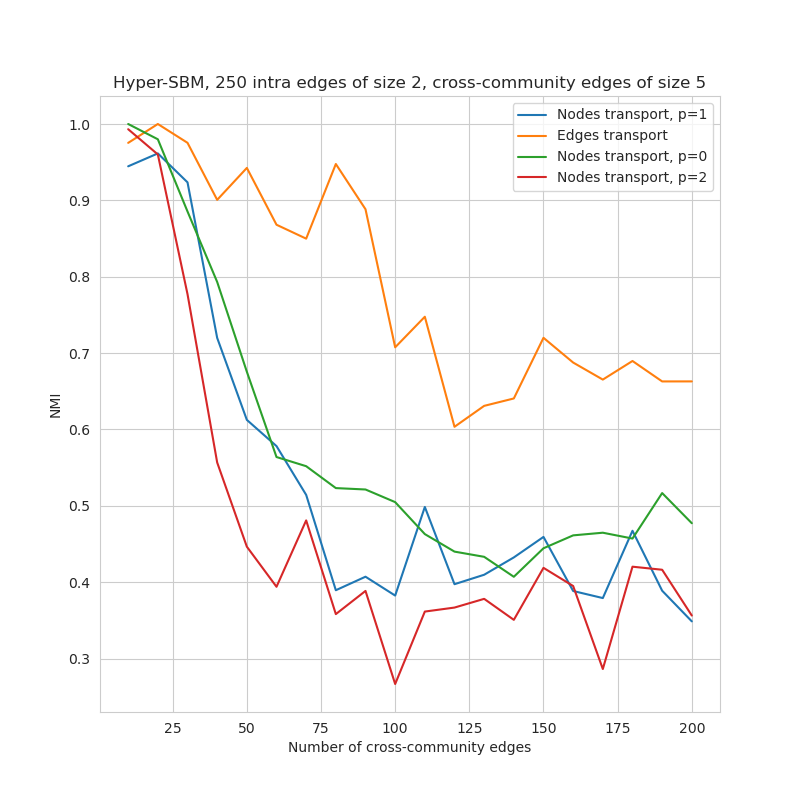}}
\subfigure[$s_{in} = 3, s_{out} = 5$]{\includegraphics[scale=0.19]{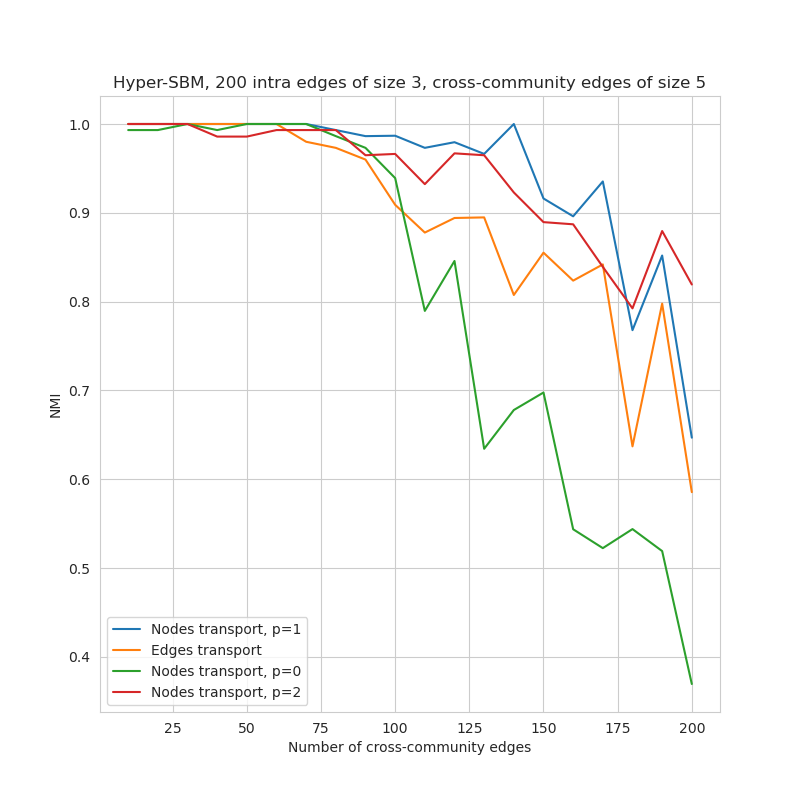}}
\subfigure[$s_{in} = 4, s_{out} = 5$]{\includegraphics[scale=0.19]{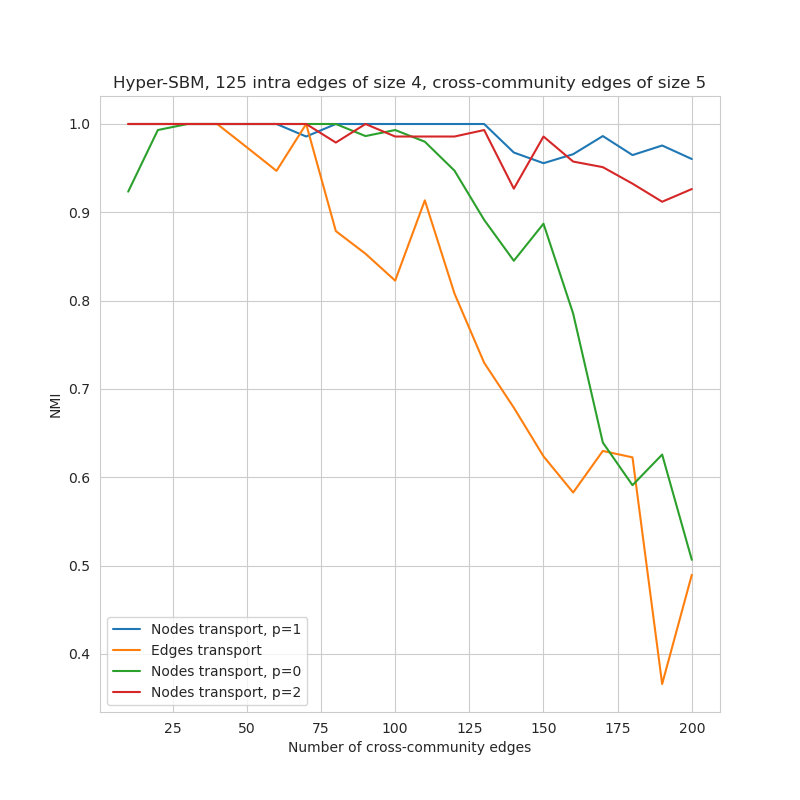}} \\
\end{center}
\caption{Study on the impact of the parameter $p$ to cluster hypergraph stochastic block models.}
\label{fig:ablation_p}
\end{figure}

\paragraph{Measure parameter $\alpha$.} 

\begin{figure}[hbtp]
\begin{center}
\subfigure[$s_{in} = 3, s_{out} = 6$]{\includegraphics[scale=0.19]{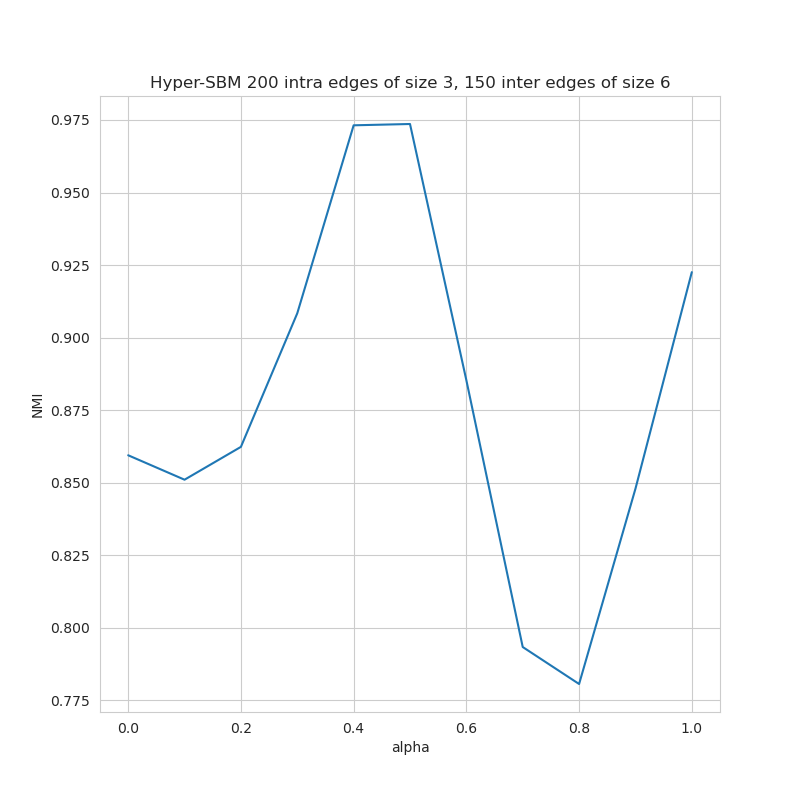}}
\subfigure[$s_{in} = 3, s_{out} = 4$]{\includegraphics[scale=0.19]{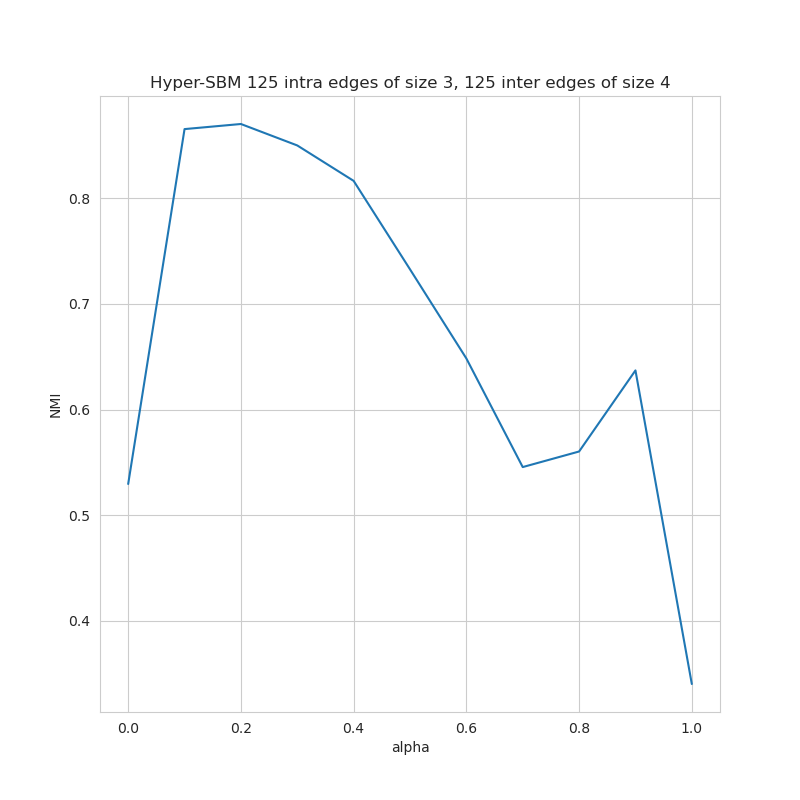}}
\subfigure[$s_{in} = 3, s_{out} = 4$]{\includegraphics[scale=0.19]{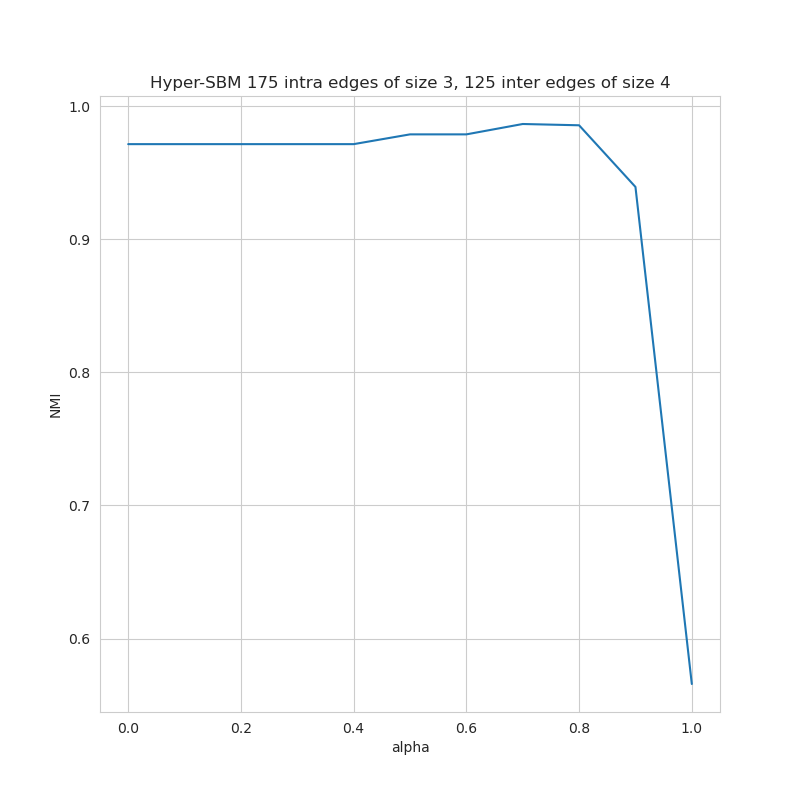}}
\\
\end{center}
\caption{Study on the impact of the parameter $\alpha$ to cluster hypergraph stochastic block models}
\label{fig:ablation_alpha}
\end{figure}

In Figure \ref{fig:ablation_alpha}, we report the NMI for a few specific critical values of $s_{in}, s_{out}, N_{in}, N_{out}$. These results are obtained for $p=1$ for the node transport. We set $Agg$ to be the maximum function and consider a uniform weighting of the clique graph. We can see that in these critical cases, the parameter $\alpha$ can play a very big impact. It is difficult to provide any generic guideline, but extreme values of $\alpha$ seem to provoke instabilities in some cases.

\paragraph{Aggregation and clique graph weighting}

\begin{figure}[hbtp]
\begin{center}
\subfigure[$s_{in} = 2, s_{out} = 5$]{\includegraphics[scale=0.19]{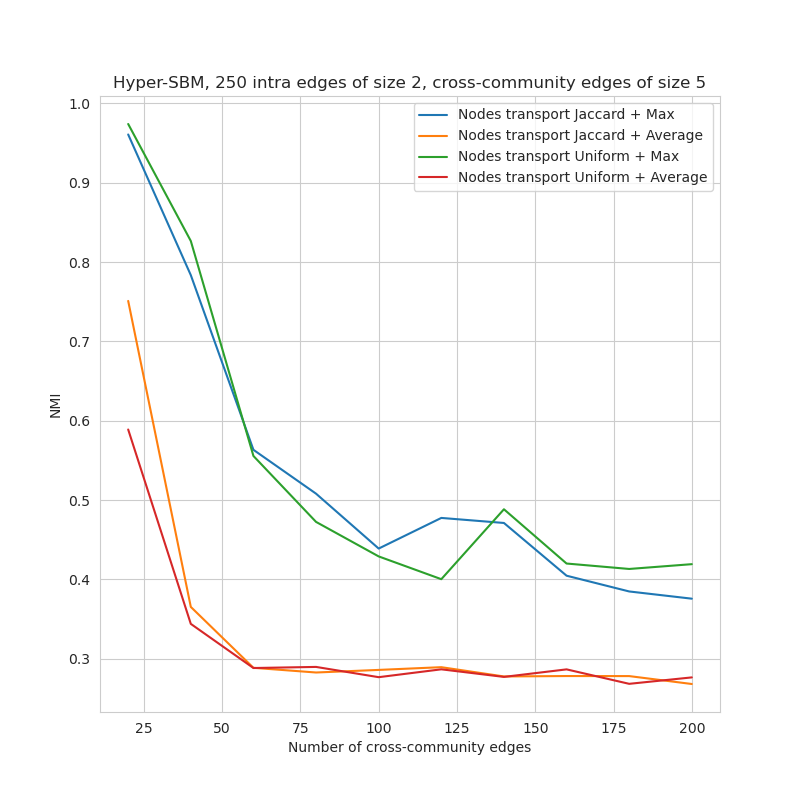}}
\subfigure[$s_{in} = 3, s_{out} = 5$]{\includegraphics[scale=0.19]{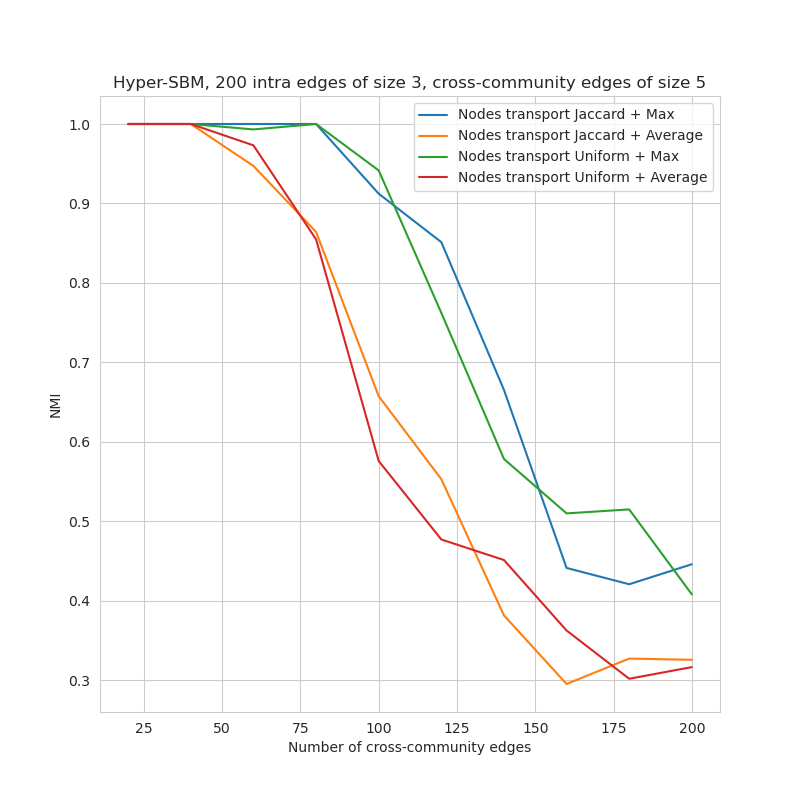}}
\subfigure[$s_{in} = 5, s_{out} = 4$]{\includegraphics[scale=0.19]{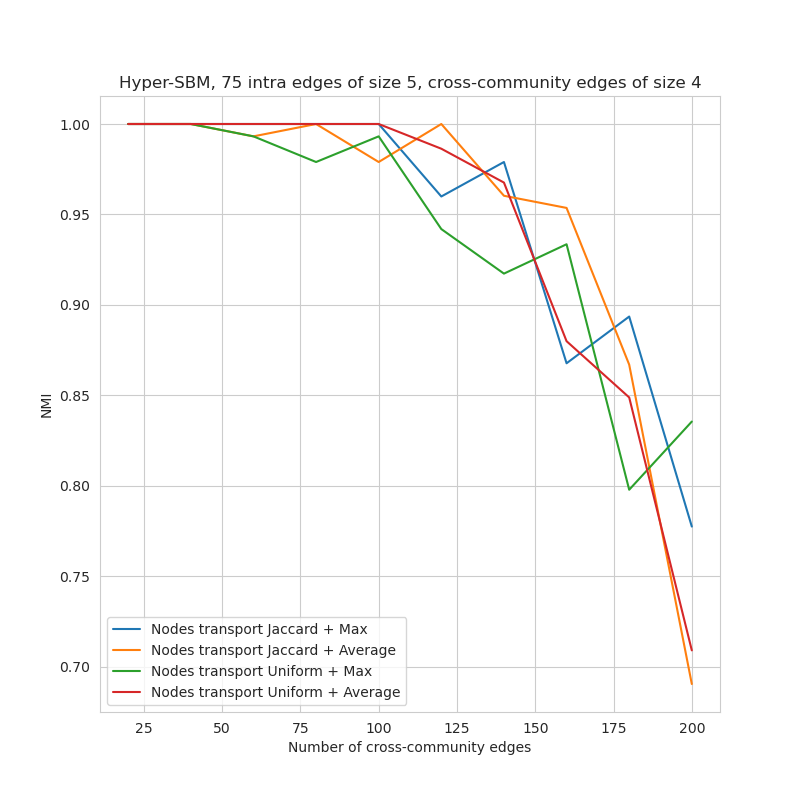}}
\\
\end{center}
\caption{Study on the impact of the clique graph weighting and the aggregation function to cluster hypergraph stochastic block models with node-Ricci flow}
\label{fig:ablation_jac_max_nodes}
\end{figure}

In Figure \ref{fig:ablation_jac_max_nodes} we report the NMI in the synthetic set-up of Section \ref{sec:synthetic} where we simultaneously compare the effect of the clique graph weighting and the curvature aggregation at the hyperedge level. We compare the four possible combinations, where the clique graph can have either a Jaccard or a uniform weighting, and $Agg$ can be either the maximum or the arithmetic average. We do not notice any significant difference between the two types of clique graph weightings. However, we observe that using the maximum pairwise curvature instead of the average yields similar to much better results and should therefore always be preferred.  

\paragraph{Aggregation and line graph weighting}

\begin{figure}[hbtp]
\begin{center}
\subfigure[$s_{in} = 2, s_{out} = 4$]{\includegraphics[scale=0.22]{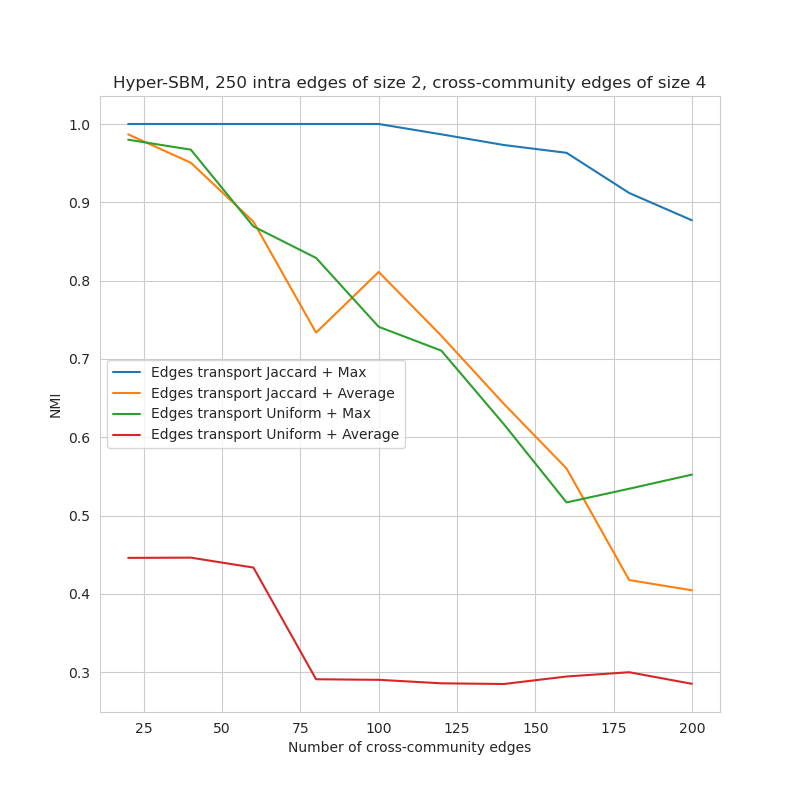}}
\subfigure[$s_{in} = 4, s_{out} = 2$]{\includegraphics[scale=0.22]{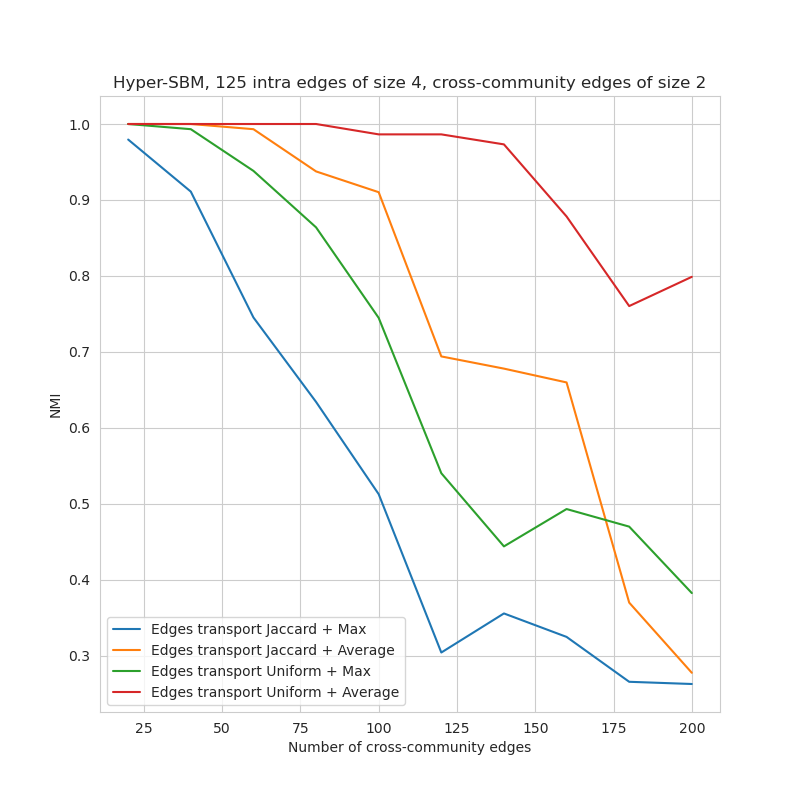}}
\subfigure[$s_{in} = 4, s_{out} = 4$]{\includegraphics[scale=0.22]{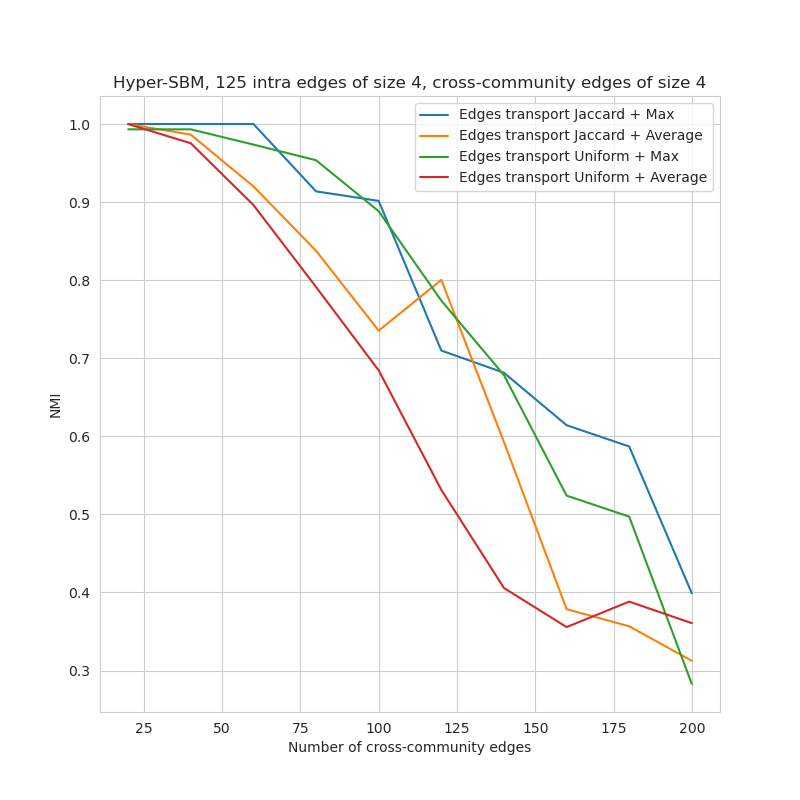}}
\\
\end{center}
\caption{Study on the impact of the line graph weighting and the aggregation function to cluster hypergraph stochastic block models with edge-Ricci flow}
\label{fig:ablation_jac_max_edges}
\end{figure}

\begin{figure}[hbtp]
\begin{center}
\subfigure[$s_{in} = 2, s_{out} = 4$]{\includegraphics[scale=0.22]{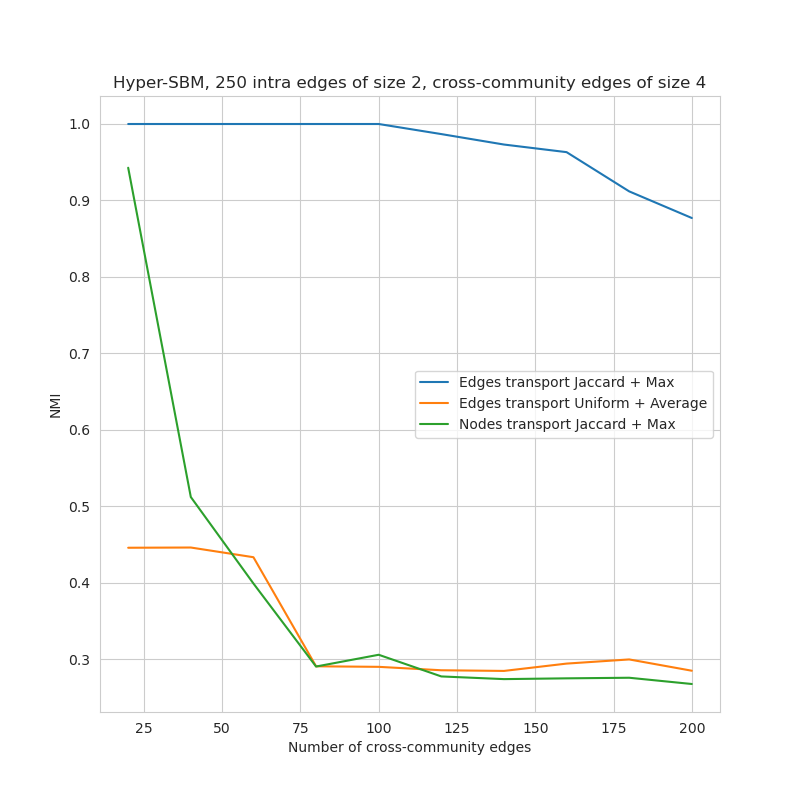}}
\subfigure[$s_{in} = 4, s_{out} = 2$]{\includegraphics[scale=0.22]{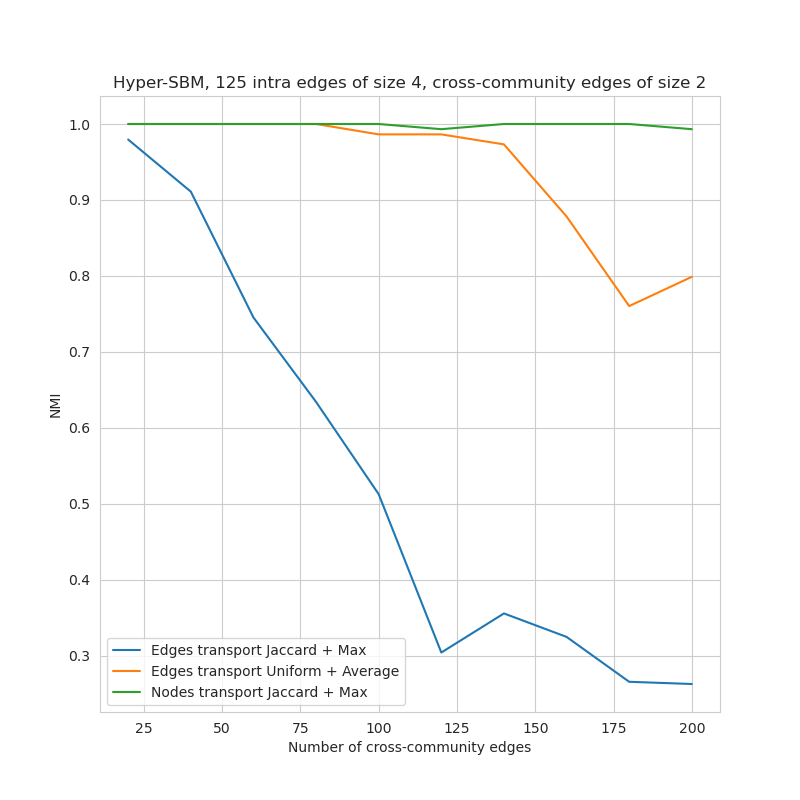}}
\subfigure[$s_{in} = 5, s_{out} = 3$]{\includegraphics[scale=0.22]{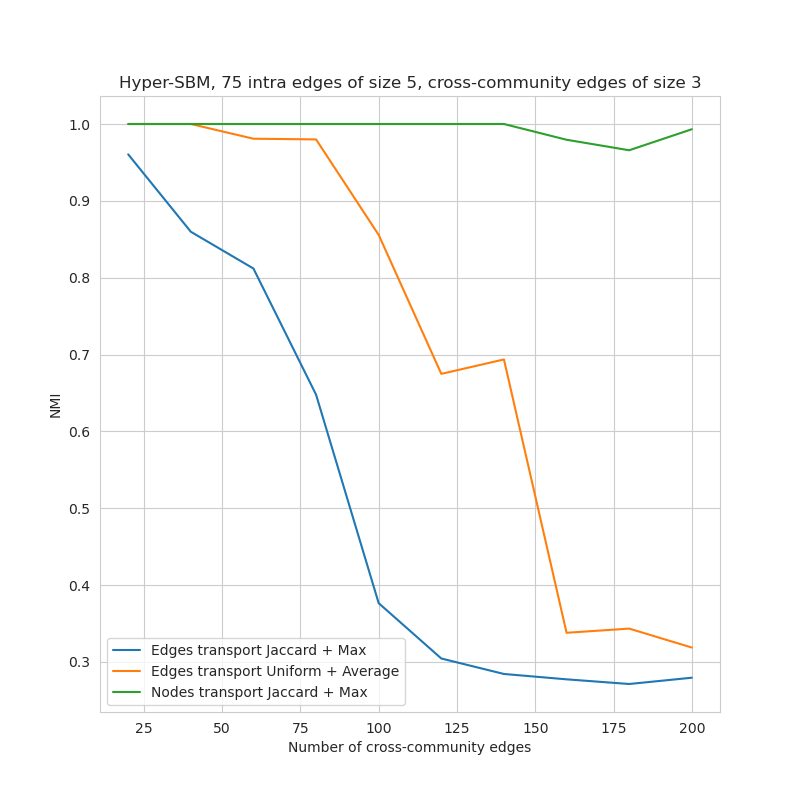}}
\\
\end{center}
\caption{Comparison of Jaccard-maximum and Uniform-average edge-Ricci flow with node-Ricci flow.}
\label{fig:ablation_comparison_NE}
\end{figure}

In Figure \ref{fig:ablation_jac_max_edges} we report the NMI in the synthetic set-up of Section \ref{sec:synthetic} where we simultaneously compare the effect of the line graph weighting and the curvature aggregation at the hyperedge level. We compare the four possible combinations, where the line graph can have either a Jaccard or a uniform weighting, and $Agg$ can be either the maximum or the arithmetic average. The differences can be very significative, in particular between the combinations Jaccard-maximum and Uniform-average. Overall, we observe that the former provides a much better clustering accuracy whenever $s_{in} \leq s_{out}$ while the latter is better when $s_{in} > s_{out}$. However, we observe in Figure \ref{fig:ablation_comparison_NE} that in this case, edge-Ricci flow with any parameter still underperforms compared to node-Ricci flow, which is particularly adapted to dealing with the situation of large intra-community hyperedges and small inter-community hyperedges. We consider a node-Ricci flow with parameters Jaccard-maximum. As a result, we claim that the parameter combination Uniform-average for edge-flow should never be preferred as it will always be beaten by either the Jaccard-maximum edge-flow or by the node-Ricci flow.

\paragraph{Joint study of every hyperparameter on a real dataset.}

We study the impact of all parameters except $\alpha$ and $\tau$ for the clustering task on the dataset \texttt{NTU2012}. The clustering accuracies are reported for the optimal $\tau^\star$. Results are reported in Tables \ref{tab:ablation_N} and \ref{tab:ablation_E}. We can see that in this real data case, The choice of the weights for the clique graph and the line graph has a negligible impact. For both types of Ricci flow, considering the maximum curvature over the average one has a tremendous impact. In addition, we can observe an improvement by considering $p=1$ over $p=0$. For $p=2$, numerical instabilities prevented the code from compiling.






\begin{table}[h!]
\begin{center}
\scalebox{0.9}{
\begin{tabular}{|c|c|c|c|} 

 \hline
 Aggregation & Clique graph weights & $p$ & NMI \\ 
 \hline
 \hline
 Max & Jaccard & 0 & 79.7 \\ 
\hline
 Max & Uniform & 0 & 79.7 \\ 
\hline
 Average & Jaccard & 0 & 69.1 \\ 
\hline
 Average & Uniform & 0 &  69.1 \\ 
\hline
 Max & Jaccard & 1 & 82.1 \\ 
\hline
 Max & Uniform & 1 & 82.5 \\ 
 \hline
  Average & Jaccard & 1 & 69.1 \\ 
\hline
 Average & Uniform & 1 & 69.1 \\ 
 \hline
\end{tabular}}
\caption{Node-Ricci flow clustering of the \texttt{NTU2012} dataset for various sets of hyperparameters}
\label{tab:ablation_N}
\end{center}
\end{table}

\begin{table}[h!]
\begin{center}
\scalebox{0.9}{
\begin{tabular}{|c|c|c|} 

 \hline
 Aggregation & Clique graph weights & NMI \\ 
 \hline
 \hline
 Max & Jaccard & 80.2 \\ 
\hline
 Max & Uniform & 80.2 \\ 
\hline
 Average & Jaccard & 69.1 \\ 
\hline
 Average & Uniform & 69.1 \\ 
\hline
\end{tabular}}
\caption{Edge-Ricci flow clustering of the \texttt{NTU2012} dataset for various sets of hyperparameters}
\label{tab:ablation_E}
\end{center}
\end{table}

\paragraph{Threshold parameter $\tau$}
We now try to assess the estimation of the $\tau^\star$ parameter providing the highest clustering accuracy. We compare the NMI obtained by trimming edges with a flow greater than $\tau^\star$ with the NMI obtained with $\tau_H$ and $\tau_C$. We report the results for the clustering of the datasets \texttt{NTU2012} and \texttt{Zoo} in Tables \ref{tab:tau_ntu} and \ref{tab:tau_zoo}. Based on the previous findings, we considered a Jaccard weighting, a hyperedge aggregation with the $\max$ function, and $p=1$ for the nodes transport. On the \texttt{NTU2012} dataset, we observe a better performance using the hypergraph notion of modularity over the clique graph modularity. The loss of accuracy for Edge-Ricci flow is milder than for Node-Ricci flow when using an approximate $\tau$ parameter. For the \texttt{Zoo} dataset, we observe a very strong difference between the two notions of modularity, as considering the hypergraph modularity yields the optimal clustering, while the modularity on the clique graph gives extremely poor results. This has to be paralleled with the results from Table \ref{tab:results} where methods relying only on the clique graph expansion provided very unsatisfactory results. 

\begin{table}[h!]
\begin{center}
\scalebox{0.9}{
\begin{tabular}{|c|c|c|c|} 

 \hline
 Type of flow & $\tau^\star$ & $\tau_H$ & $\tau_C$ \\ 
 \hline
 \hline
Nodes & 82.1 & 76.3 & 72.3 \\ 
\hline
Edges & 80.2 & 77.6 & 75.2 \\ 
\hline
\end{tabular}}
\caption{NMI for various choices of $\tau$ on the \texttt{NTU2012} dataset}
\label{tab:tau_ntu}
\end{center}
\end{table}

\begin{table}[h!]
\begin{center}
\scalebox{0.9}{
\begin{tabular}{|c|c|c|c|} 

 \hline
 Type of flow & $\tau^\star$ & $\tau_H$ & $\tau_C$ \\ 
 \hline
 \hline
Nodes & 96.2 & 96.2 & 18.3 \\ 
\hline
Edges & 100 & 100 & 18.3 \\ 
\hline
\end{tabular}}
\caption{NMI for various choices of $\tau$ on the \texttt{Zoo} dataset}
\label{tab:tau_zoo}
\end{center}
\end{table}

\end{document}